%% file: ieee_tnnls.tex
\documentclass[journal, twoside]{IEEEtran}

\usepackage{times}
\usepackage{epsfig}
\usepackage{graphicx}
\usepackage{amsmath}
\usepackage{amssymb}
\usepackage[numbers,sort]{natbib}

\usepackage[utf8]{inputenc}
\usepackage{amsfonts}    
\usepackage{nicefrac}       
\usepackage{microtype}      
\usepackage{multirow}
\usepackage{booktabs}
\usepackage{amsmath}
\usepackage{mathtools}
\usepackage{amssymb}
\usepackage{enumitem}
\usepackage{tabularx}
\usepackage{adjustbox}
\usepackage{threeparttable}
\usepackage{subfig}
\usepackage{algorithm}
\usepackage{algorithmic}
\usepackage{etoolbox}
\usepackage{lipsum}
\usepackage{cuted}
\usepackage{capt-of}
\usepackage{dsfont}

\usepackage[switch]{lineno}
\usepackage{color}
\usepackage{xcolor}
\usepackage{colortbl}
\usepackage{placeins}
\usepackage{kotex}

\usepackage{scalerel}
\usepackage{tikz}
\usetikzlibrary{svg.path}

\definecolor{orcidlogocol}{HTML}{A6CE39}
\tikzset{
  orcidlogo/.pic={
    \fill[orcidlogocol] svg{M256,128c0,70.7-57.3,128-128,128C57.3,256,0,198.7,0,128C0,57.3,57.3,0,128,0C198.7,0,256,57.3,256,128z};
    \fill[white] svg{M86.3,186.2H70.9V79.1h15.4v48.4V186.2z}
                 svg{M108.9,79.1h41.6c39.6,0,57,28.3,57,53.6c0,27.5-21.5,53.6-56.8,53.6h-41.8V79.1z M124.3,172.4h24.5c34.9,0,42.9-26.5,42.9-39.7c0-21.5-13.7-39.7-43.7-39.7h-23.7V172.4z}
                 svg{M88.7,56.8c0,5.5-4.5,10.1-10.1,10.1c-5.6,0-10.1-4.6-10.1-10.1c0-5.6,4.5-10.1,10.1-10.1C84.2,46.7,88.7,51.3,88.7,56.8z};
  }
}

\newcommand\orcidicon[1]{\href{https://orcid.org/#1}{\mbox{\scalerel*{
\begin{tikzpicture}[yscale=-1,transform shape]
\pic{orcidlogo};
\end{tikzpicture}
}{|}}}}

\newcommand{\figref}[1]{Fig.~\ref{#1}}
\newcommand{\tabref}[1]{Table~\ref{#1}}

\newcommand{\secref}[1]{Sec.~\ref{#1}}
\newcommand{\eg}{\textit{e.g.}}
\newcommand{\ie}{\textit{i.e.}}
\newcommand*\rot{\rotatebox{90}}

\definecolor{Gray}{gray}{0.9}
\definecolor{myred}{rgb}{0.8, 0.1, 0.1}
\definecolor{mygreen}{rgb}{0.2, 0.7, 0.1}
\usepackage[pagebackref=true,breaklinks=true,colorlinks,linkcolor={myred},citecolor={mygreen},bookmarks=false]{hyperref}

\title{
    Tackling the Challenges in Scene Graph Generation \\
    with Local-to-Global Interactions
}

\author{
    Sangmin~Woo$^{\textsuperscript{\orcidicon{0000-0003-4451-9675}}}$,~\IEEEmembership{Student~Member,~IEEE,}
    Junhyug~Noh$^{\textsuperscript{\orcidicon{0000-0003-1239-8178}}}$,~\IEEEmembership{Member,~IEEE,}
    and~Kangil~Kim$^{\textsuperscript{\orcidicon{0000-0003-3220-6401}}}$,~\IEEEmembership{Member,~IEEE}
    \thanks{
    Manuscript received June 17, 2021; revised January 11, 2022; accepted March 12, 2022.
    This work was supported in part by the National Research Foundation of Korea (NRF) grant funded by the Korea government (MSIT) (2019R1A2C1091077), and in part by the Institute of Information \& communications Technology Planning \& Evaluation (IITP) grant funded by the Korea government (MSIT) (No. 2019-0-01842, Artificial Intelligence Graduate School Program (GIST)). \textit{(Corresponding author: Kangil Kim)}}
    \thanks{Sangmin Woo is with the School of Electrical Engineering, Korea Advanced Institute of Science and Technology, Daejeon 34141, Korea. This work was done when he was an M.S. student at GIST (email: smwoo95@kaist.ac.kr).}
    \thanks{Junhyug~Noh is with the Computational Engineering Division, Lawrence Livermore National Laboratory, CA 94550, United States (email: noh1@llnl.gov).}
    \thanks{Kangil Kim is with the School of Electrical Engineering and Computer Science and the AI Graduate School, Gwangju Institute of Science and Technology, Gwangju 61005, Korea (email: kangil.kim.01@gmail.com).}
    \thanks{This version is accepted to IEEE Transactions on Neural Network and Learning Systems. DOI of the published version is 10.1109/TNNLS.2022.3159990.}
    \thanks{© 2022 IEEE. Personal use of this material is permitted. Permission from IEEE must be obtained for all other uses, in any current or future media, including reprinting/republishing this material for advertising or promotional purposes, creating new collective works, for resale or redistribution to servers or lists, or reuse of any copyrighted component of this work in other works.}
}

\begin{document}
    \maketitle
    \begin{strip}
        \centering
        \vspace{-44mm}
        \includegraphics[width=\linewidth]{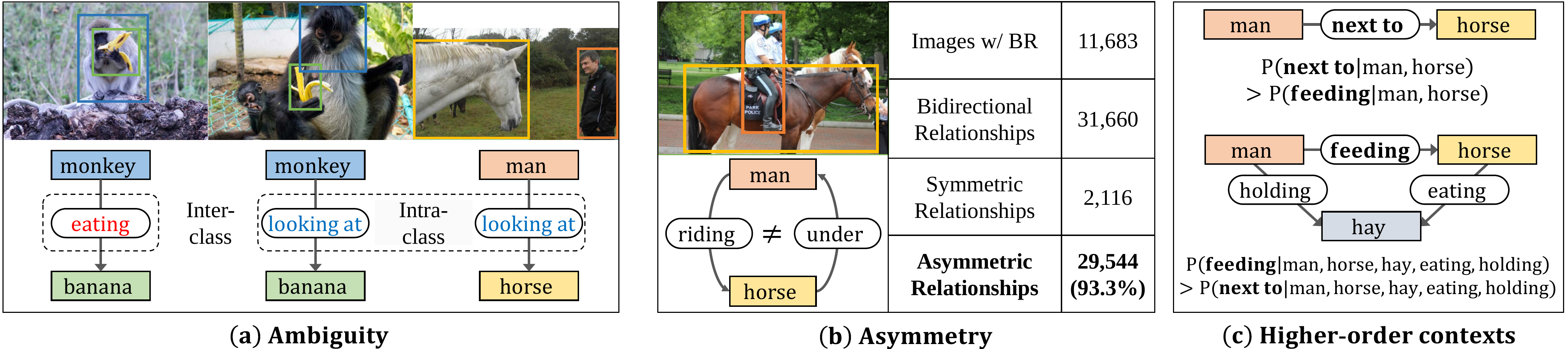}
        \captionsetup{font=footnotesize}
        \captionof{figure}{
            \textbf{Challenges in Scene Graph Generation.}
            (a) \textbf{Ambiguity}: different predicates may be visually similar (first two), and the same predicate may not be visually similar (last two).
            (b) \textbf{Asymmetry}: relationships have direction, and those relationships in the opposite direction are mostly different (asymmetric).
            (c) \textbf{Higher-order contexts}: the other components in the scene serve as contexts while predicting the relationship. Data statistics are based on the Visual Genome dataset~\cite{krishna2017visual}.
        }
        \label{fig:Motivation}
    \end{strip}

    \begin{abstract}
        In this work, we seek new insights into the underlying challenges of the Scene Graph Generation (SGG) task. Quantitative and qualitative analysis of the Visual Genome dataset implies --
        1) \textbf{Ambiguity}: even if inter-object relationship contains the same object (or predicate), they may not be visually or semantically similar,
        2) \textbf{Asymmetry}: despite the nature of the relationship that embodied the direction, it was not well addressed in previous studies, and
        3) \textbf{Higher-order contexts}: leveraging the identities of certain graph elements can help to generate accurate scene graphs.
        Motivated by the analysis, we design a novel SGG framework, Local-to-Global Interaction Networks (LOGIN).
        Locally, interactions extract the essence between three instances of subject, object, and background, while baking direction awareness into the network by explicitly constraining the input order of subject and object.
        Globally, interactions encode the contexts between every graph component (\ie, nodes and edges).
        Finally, Attract \& Repel loss is utilized to fine-tune the distribution of predicate embeddings.
        By design, our framework enables predicting the scene graph in a bottom-up manner, leveraging the possible complementariness.
        To quantify how much LOGIN is aware of relational direction, a new diagnostic task called Bidirectional Relationship Classification (BRC) is also proposed.
        Experimental results demonstrate that LOGIN can successfully distinguish relational direction than existing methods (in BRC task), while showing state-of-the-art results on the Visual Genome benchmark (in SGG task).
    \end{abstract}
    
    
    \input{1.introduction}
    \input{2.related_work}
    \input{3.motivation}
    \input{4.method}
    \input{5.experiments}
    \input{6.conclusion}
    \input{7.references}
    \input{8.biography}
\end{document}

%% file: 1.introduction.tex
\section{Introduction}
    \IEEEPARstart{T}{o} understand a scene, inferring underlying properties such as the relationship between entities (In this work, we use the term “entity” to describe individual detected object instances to distinguish them from “object” in the semantic sense) is just as important as observing explicit information about what and where entities are. However, most state-of-the-art visual recognition models focus on detecting individual entities in isolation~\cite{girshick2015fast, he2017mask, long2015fully, redmon2016you, simonyan2014very}, and they are still far from reaching the goal of capturing their relationships. In an effort to incorporate the relational reasoning ability into the model, a scene graph representation -- a structured description that captures semantic summaries of entities and their relationships -- has been presented recently~\cite{johnson2015image}. Since then, a number of works have proposed deep network-based approaches for generating the scene graphs, confirming its importance to the field~\cite{teney2017graph, yao2018exploring, johnson2018image, ma2018attend, yang2019auto, ashual2019specifying, li2020context}.
    While scene graph representation holds tremendous promise, extracting scene graphs from images is known to be challenging.
    
    In~\secref{section:challenges}, we first explore what the fundamental challenges of the task are:
    \begin{enumerate}
    \item \textbf{Ambiguity}:
        We postulate the main cause of ambiguity is due to high intra- and low inter-class variability of predicates.
        Although there is little visual difference between the images, the predicates can be different, and vice versa (\figref{fig:Motivation} (a)).
        Therefore, the model should be aware of the inconsistency between the visual and the actual predicate. In other words, we require a model that can recognize a subtle visual difference to differentiate the predicates and learn that the predicates can be the same in a completely different visual context.
    \item \textbf{Asymmetry}:
        By nature, a relationship has a direction. Also, we can always define relationships in both directions. Nevertheless, we see that understanding the relational direction has not been well established in previous studies, and there is a lack of consideration on how to effectively address it.
        In this work, we are particularly interested in bidirectional relationships with asymmetry (\figref{fig:Motivation} (b)). To analyze how much the model understands relational direction, we introduce a new diagnostic task called \textit{Bidirectional Relationship Classification (BRC)}.
    \item \textbf{Higher-order contexts}:
        Often relations need to be considered with the contextual dependency of the whole scene beyond being defined as a pair-wise relation. Suppose there is a horse close to a man (\figref{fig:Motivation} (c)). Without any other clue, one might say that their relationship is \texttt{"next to"}. However, if the presence of other entities and relationships (\eg, \texttt{hay, eating}) is known, the relationship between the horse and the man is more likely to be \texttt{"feeding"}.
        To examine the benefits of higher-order contexts, we quantitatively analyze the amount of information gain given each graph component.
    \end{enumerate}
    
    In~\secref{section:login}, with the aforementioned issues in mind, we present a novel framework, \textbf{Lo}cal-to-\textbf{G}lobal \textbf{I}nteraction \textbf{N}etworks \textbf{(LOGIN)}.
    First, LOGIN highlights informative representation between three entity-level instances by weighing how much each pair-wise interaction contributes to relational representation.
    Second, direction awareness is baked into the model by fusing feature instances in a constrained order (\eg, subject precedes object).
    Third, LOGIN considers interaction between every scene graph element. The informative contexts essential to accurately predict each graph component are propagated to every graph component.
    Last but not least, we introduce Attract \& Repel Loss, which effectively scales the variability within and between classes making the model robust against ambiguity. We explain this in more detail in~\secref{section:loss}.
    By design, LOGIN effectively leverages the complementariness of entity-level interactions and graph-level interactions.
    
    Finally, in~\secref{section:experiments}, we evaluate our final model on both the Visual Genome benchmark and the BRC task.
    By ablating each network design, we observe that all design principles cooperate in generating visually grounded scene graphs.
    Unifying all design principles into a single framework, LOGIN achieved state-of-the-art results on the Visual Genome benchmark while outperforming competing approaches by a comfortable margin on the BRC task.
    
    Our contributions can be summarized as follows:
    \begin{itemize}
        \item Through quantitative and qualitative analysis on the Visual Genome dataset, we identify fundamental challenges in the SGG task: 1) Ambiguity, 2) Asymmetry, 3) Higher-order contexts.
        \item We design a novel framework, Local-to-Global Interaction Networks (LOGIN), to address the aforementioned issues, which achieved competitive results against state-of-the-arts on the Visual Genome benchmark.
        \item To quantify and concretely see how well the model understands the relational direction, we introduce a new Bidirectional Relationship Classification (BRC) task. Here, LOGIN significantly outperformed state-of-the-art by a 6\% of mean performance gain.
    \end{itemize}

%% file: 2.related_work.tex
\section{Related Work}
    Numerous works have actively studied the task of recognizing entities and their relationships in various forms. 
    This includes entity localization from natural language expressions~\cite{hu2017modeling}, human-entity interactions~\cite{gkioxari2018detecting,chao2018learning, li2019transferable}, or the more general tasks of visual relationship detection~\cite{lu2016visual,zhang2017visual,li2017vip,plummer2017phrase,zhang2017ppr,dai2017detecting,liang2017deep, mi2020hierarchical}, and scene graph generation~\cite{xu2017scene,zellers2018neural,woo2018linknet,yang2018graph,li2018factorizable,qi2019attentive,chen2019knowledge,tang2019learning,chen2019counterfactual, zhang2019graphical, zareian2020weakly,tang2020unbiased,zareian2020bridging,wang2020sketching}.
    
    Among them, the scene graph generation has recently drawn much attention. 
    The challenging and open-ended nature of the task lends itself to a variety of diverse methods.
    For example, 
    refining entity and predicate labels using iterative message passing~\cite{xu2017scene};
    staging the generation process in three-step based on the observation that entity labels are highly predictive of predicate labels~\cite{zellers2018neural};
    explicitly modeling inter-dependency among entire entities using bi-linear pooling~\cite{woo2018linknet};
    leveraging the idea of proposal network~\cite{ren2015faster} and graph convolution~\cite{kipf2017semi} jointly~\cite{yang2018graph};
    combining both visual and linguistic features to exploit linguistic analogies~\cite{qi2019attentive}; 
    using statistical correlations between entity pairs and their relationships to regularize semantic space~\cite{chen2019knowledge};
    presenting a multi-agent policy gradient method to replace standard cross-entropy loss and maximize a graph-level metric~\cite{chen2019counterfactual};
    disentangles entity and predicate recognition, enabling sub-quadratic performance~\cite{zareian2020weakly}.
    
    We shed light on three underlying challenges that have not been dealt with in-depth in previous studies: 1) Ambiguity, 2) Asymmetry, and 3) Higher-order contexts (see~\figref{fig:Motivation}).
    Similar to ours, the ambiguity issue has been addressed in~\cite{zhang2019graphical}, which is a proximal relationship ambiguity arising from multiple subject-object pairs being gathered nearby. On the other hand, we interpret it differently as visual and semantic ambiguity caused by high intra-class variance and low inter-class variance.
    To the best of our knowledge, the asymmetry issue in SGG is explicitly and importantly addressed for the first time in this work. We believe that some works~\cite{zhang2017visual, yang2018shuffle, zellers2018neural, zhang2019graphical, tang2019learning, tang2020unbiased} can also cope with relational directions, albeit they do not suggest an effective method.
    Higher-order context problems have been addressed a lot in previous SGG studies~\cite{xu2017scene, zellers2018neural, woo2018linknet, yang2018graph}, but most focus on context utilization aspects.
    We rather take a slightly different approach to find the answer to the question, \textit{``Are we fully exploiting all the information available?"}.
    To this end, we examine the predictability of identity according to given graph components and then apply the most promising way we find to the context propagation step.
    In summary, we design LOGIN, an integrated framework based on the analysis, to tackle the challenges simultaneously.

%% file: 3.motivation.tex
\section{Identifying Challenges in Scene Graph Generation}
    \label{section:challenges}
    This section seeks quantitative insights on the underlying challenges of the SGG task by analyzing the Visual Genome dataset.
    In particular, 1) \textbf{Ambiguity} (\secref{section:ambiguity}): how intra- and inter-class variance hinder clearly differentiating the predicate class boundary,
    2) \textbf{Asymmetry} (\secref{section:asymmetry}): how has relational direction been overlooked, and how can direction awareness be quantified,
    and 3) \textbf{Higher-order contexts} (\secref{section:contexts}): what higher-order context should be considered to predict the identity of each graph element.
    Motivated by our findings, we design the LOGIN to better integrate local and global contexts, which will be described in more detail in~\secref{section:login}.
    
    \subsection{Ambiguity}
    \label{section:ambiguity}
        \begin{figure}[t!]
            \centering
            \includegraphics[width=\linewidth]{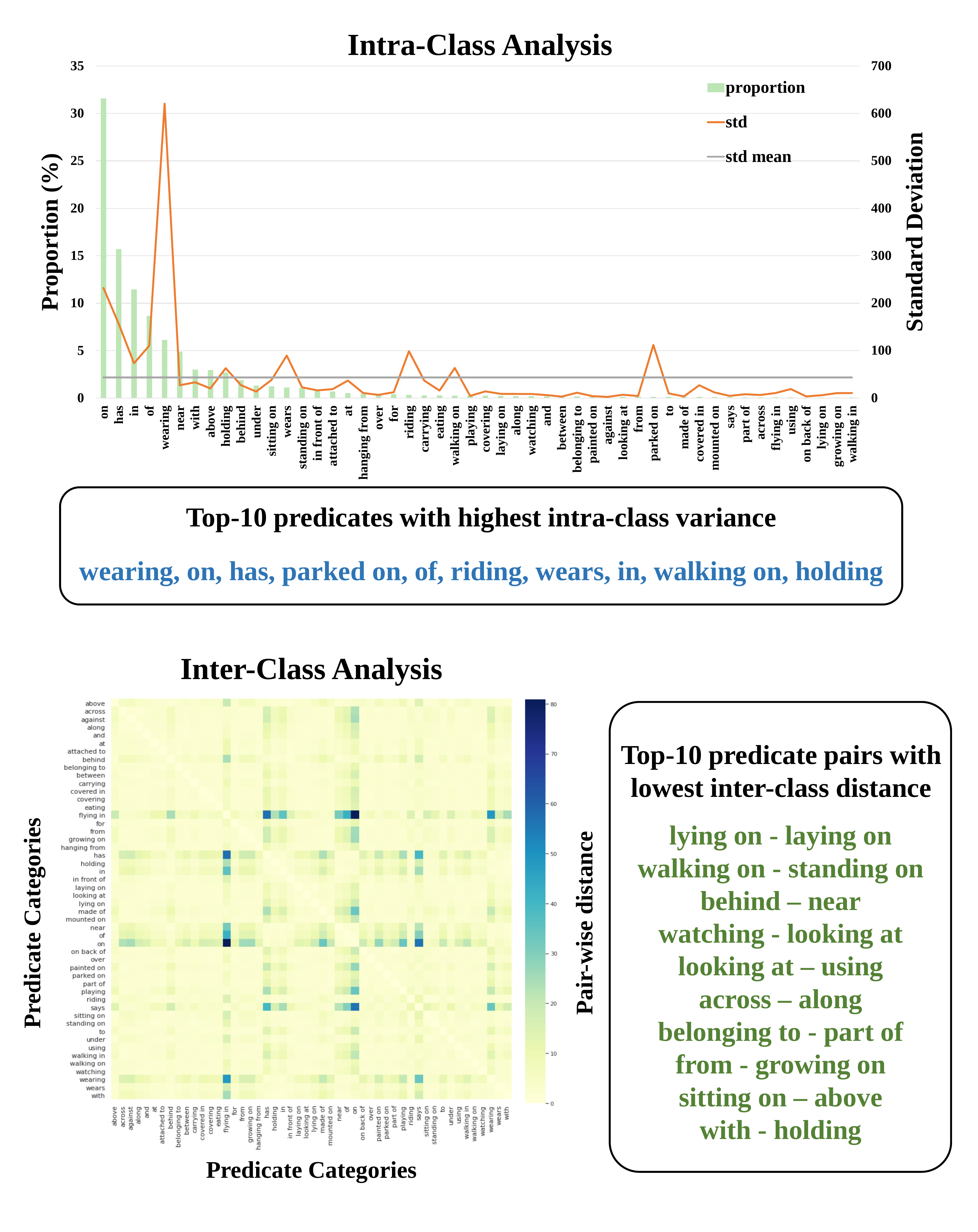}
            \captionsetup{font=footnotesize}
            \caption{
                \textbf{Intra- and inter-class analysis on predicates.} \textbf{(top)} predicate labels are arranged in the order of proportion (green bar). More frequent predicates tend to have higher intra-class variance (orange line). \textbf{(bottom)} Each block represents the degree to which predicate-predicate pair share the same entity pairs in color -- the more overlapping entity pairs, the brighter the block color. Except for a few predicate pairs (\eg, on-flying in), most predicate pairs have low inter-class distance -- the closer the brighter. Axis labels best viewed zoomed in on screen.
            }
            \label{fig:ambiguity}
        \end{figure}
        To gain insight into the Visual Genome scene graphs, we first examine the intra-class variance and inter-class distance within and in between predicate categories. Specifically, we take a close look at label statistics (\eg, subject-object-predicate co-occurrence).
        Intra-class variance within $i$th predicate can be calculated as:
        \begin{eqnarray}
            \begin{aligned}
                {var}_{intra}(i) = {1 \over N^2} \sum_{k=1}^{N^2} (f_{ik} - \mu_i)^2\,.
            \end{aligned}
            \label{eq:1}
        \end{eqnarray}
        The inter-class distance between $i$th and $j$th predicate is normalized by the co-occurrence frequency:
        \begin{eqnarray}
            \begin{aligned}
                {dist}_{inter}(i,j) = {{\sum_{i=1}^{M} \sum_{j=1}^{M} \sum_{k=1}^{N^2} |f_{ik} - f_{jk}|} \over \sqrt{\sum_{i=1}^{M} f_i} \sqrt{\sum_{j=1}^{M} f_j} }\,,
            \end{aligned}
            \label{eq:2}
        \end{eqnarray}
        where $N$ and $M$ are the number of entities and predicates respectively. $f_{ik}$ denotes the co-occurrence frequency of $i$th predicate and $k$th entity pairing, and ${\mu}_i$ denotes the mean value.
        
        The results are depicted in \figref{fig:ambiguity}. From the figure, it can be observed that frequently occurring predicates tend to have high intra-class variance, which implies the dominant predicates can be used in various contexts repeatedly (\ie, even the same predicate can pair with various entity pair candidates).
        In contrast, most of the predicate-predicate pairs have similar subject-object co-occurrence distribution. In this case, even if subject-object identity is known, it becomes difficult to predict the predicate (\ie, predicates in different categories can often pair with the same entity pair).
        
        In summary, even the same predicates may not be similar visually (see \figref{fig:Motivation}) nor semantically. Accordingly, solving these ambiguity issues could play a key role in generating accurate scene graphs.
    
    \subsection{Asymmetry}
    \label{section:asymmetry}
        \begin{figure}[t]
            \centering
            \includegraphics[width=\linewidth]{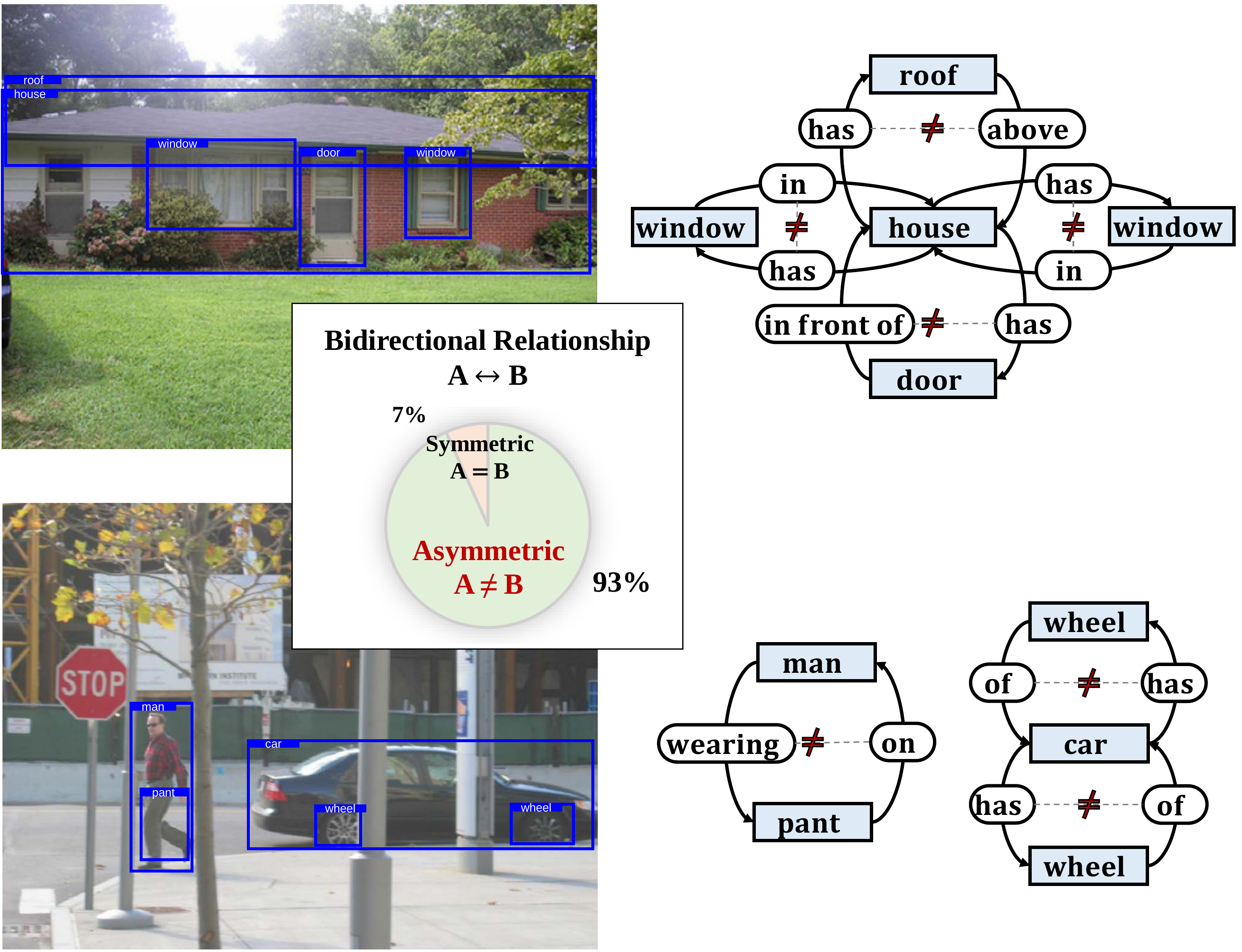}
            \captionsetup{font=footnotesize}
            \caption{
                \textbf{Examples of bidirectional relationships in Visual Genome dataset.}
                \textbf{(left)} Input images with ground-truth bounding boxes \textbf{(right)} corresponding bidirectional relationship scene graph. nodes and edges are colored in blue and green respectively. As can be seen in the figure, most bidirectional relationships are asymmetric.
            }
            \label{fig:bidirectional_examples}
        \end{figure}
        
        \paragraph{Bidirectional Relationships}
            Given two different entities $A$ and $B$, if the both $A \xrightarrow{\alpha} B$ and $B \xrightarrow{\beta} A$ relationships are defined, we consider those relationships as \emph{bidirectional relationships}. Among them, if $\alpha \neq \beta$, we denote as \emph{asymmetric relationships}, otherwise as \emph{symmetric relationships}. Examples of bidirectional relationships are shown in~\figref{fig:bidirectional_examples}.
            
            Among the total 108,073 images in the Visual Genome Dataset~\cite{krishna2017visual}, 11,683 images contain 31,660 bidirectional relationships, which can be break down into 29,544 asymmetric relationships and 2,116 symmetric relationships (see~\figref{fig:Motivation} (b)). Since the majority of relationships are asymmetric ($\sim$ 93.3\%), modeling the relational direction (with regard to the entity orderings) is crucial. 
            
        \paragraph{Modeling Relational Direction}
            One straightforward approach to obtain a visual representation of a predicate is to use a union appearance feature~\footnote{A union appearance feature is pooled from the RoI feature that tightly encompasses two (subject and object) entities.} directly, which is the form used by many previous works~\cite{xu2017scene, lu2016visual, dai2017detecting, yang2018graph, qi2019attentive, chen2019knowledge, chen2019counterfactual, li2017scene, li2018factorizable, tang2019learning}.
            Using only the union feature is straightforward and reflects the holistic representation, but it entails a fatal problem.
            For example, even the position of two entities is reversed, the union feature remains the same thus cannot embody directionality without the assistance of external features (\eg, spatial coordinates, contexts).
            This weakness is more pronounced when predicting relationships in opposite directions at the same time.
        
        \paragraph{Diagnosis of Direction Awareness}
            It is common sense that all relationships have directions and can always be defined in both directions. However, since most of the relationships that make up the Visual Genome dataset are uni-directional, a rigorous analysis on the direction awareness of model is limited. In other words, good performance can be achieved in the Visual Genome dataset without due consideration of the direction awareness.
            
            To this end, we introduce a new diagnostic task called \emph{Bidirectional Relationship Classification (BRC)} to quantify and concretely see how well the model understands the relational direction. The task is solely based on the collected images containing bidirectional relationships. Therefore, in all cases, good performance can only be achieved by understanding the direction.
            
            What we want to observe in the BRC benchmark is how much the model understands the direction of the relationship. Among the three common criteria for evaluating performance in SGG, \textit{SGGen} includes not only predicate prediction but also object localization and object class prediction, making it difficult to evaluate direction prediction intensively. Likewise, since \textit{SGCls} includes class prediction of objects, it may be difficult to strictly verify the directional understanding. Therefore, we adopt \textit{PredCls} evaluation criterion that measures only predicate predictions.
    
    \subsection{Higher-Order Contexts}
    \label{section:contexts}
        \begin{figure}
            \centering
            \includegraphics[width=0.9\linewidth]{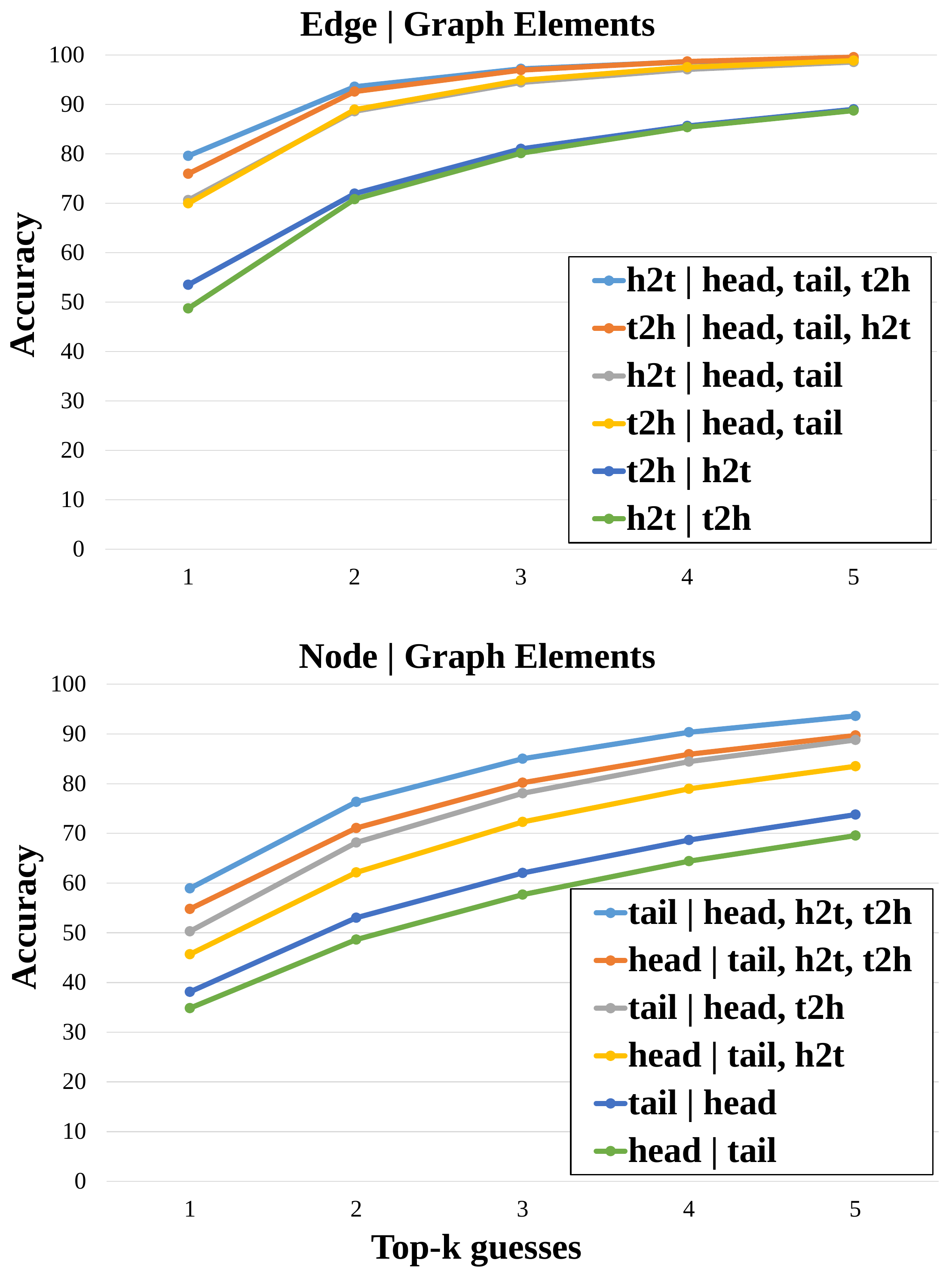}
            \captionsetup{font=footnotesize}
            \caption{
                \textbf{How much information does each graph component contain?} The figures show the likelihood of guessing the label of target element given the identity of neighboring graph components -- head, tail, and edge. Guesses were made by looking up the empirical distribution over label statistics in the training set (\eg, top-k frequent classes given graph elements). \textbf{h2t} refers to the edge from the head node to the tail node, and \textbf{t2h} refers to the edge of the opposite direction.
            }
            \label{fig:guess}
        \end{figure}
        
        To investigate the benefits of higher-order context, we measure how much information is gained given the identity of different scene graph elements.
        Motivated by~\cite{zellers2018neural}, we plot the likelihood of guessing labels of target element given labels of other graph elements in \figref{fig:guess}.
        In addition to node-conditioned guessing performed on prior work~\cite{zellers2018neural}, we further analyze the predictability improvement given the edge identity.
        To disentangle the significance of semantic knowledge from image cues (\eg, appearance, spatial), no image features are used and are guessed using only label statistics (\ie, subject-object-predicate co-occurrence).
        A higher curve implies that given graph elements are more decisive in guessing the target element.
        
        In the case of edge, it is greatly affected by the identity of neighboring nodes, consistent with our intuition.
        What is more noteworthy here is that even only one edge in the opposite direction is known, nearly 90\% accuracy can be achieved within just five guesses. It can also provide complementary information in determining the identity of the target edge when given with the neighboring nodes' identity.
        
        In the case of node, it has less correlation with adjacent graph elements than edges. However, as shown in the figure, a significant amount of information can be obtained whenever the identity of adjacent graph elements is known one by one. This fact motivates the use of as much information as possible to correctly recognize the identity of each element.
        
        To sum up, we see that both node and edge can most effectively exploit inductive bias when utilizing all the identities of adjacent graph elements.

%% file: 4.method.tex
\begin{figure*}[t!]
    \centering
    \includegraphics[width=\linewidth]{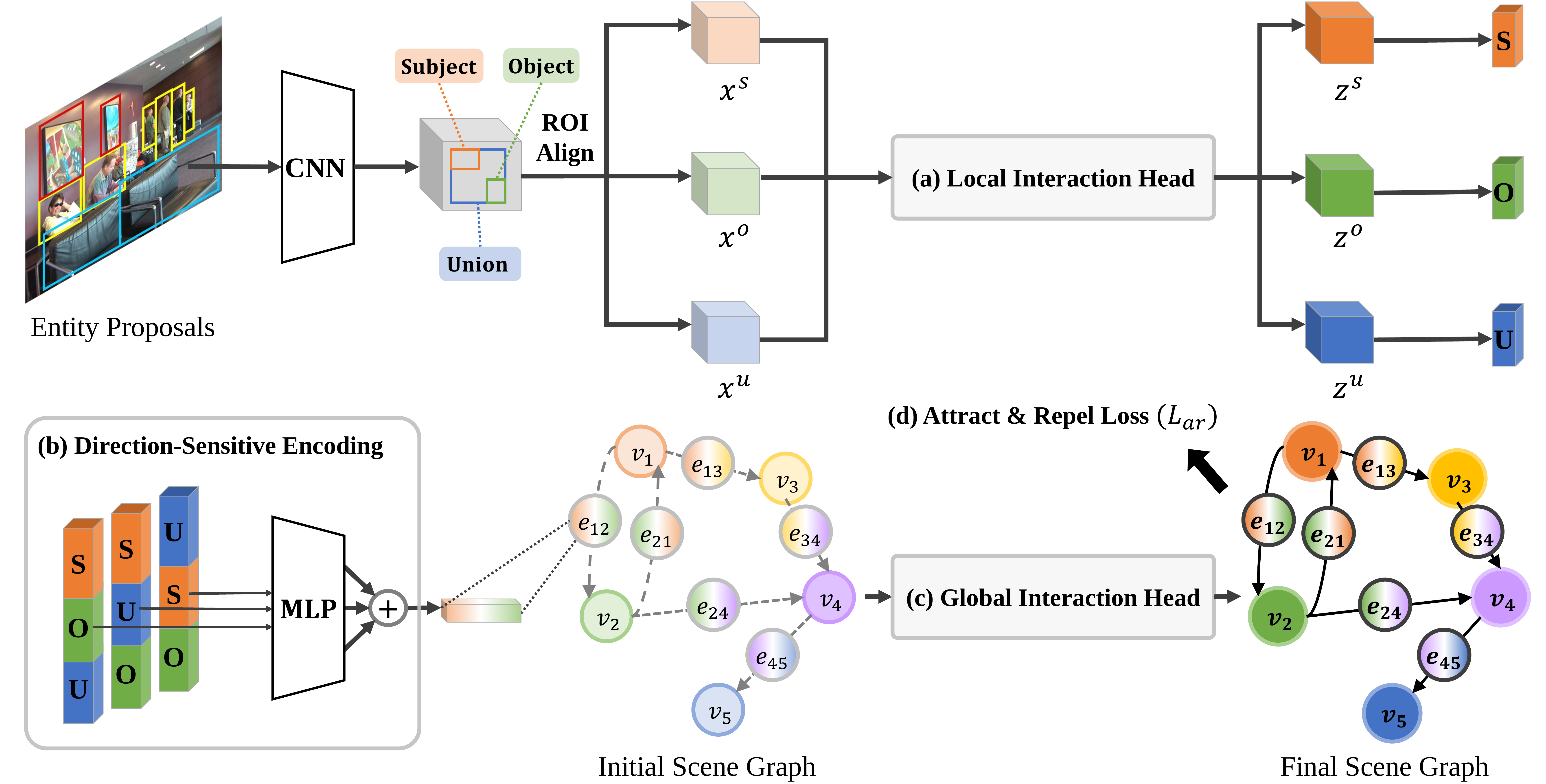}
    \captionsetup{font=footnotesize}
    \caption{
        \textbf{A high-level overview of LOGIN.}
        (a) \textbf{Local Interaction Head} (\secref{section:local}): locally, interactions extract the essence between three instances -- subject, object, and background. 
        (b) \textbf{Direction-Sensitive Encoding} (\secref{section:direction}): model become aware of relational direction by constraining the input order. 
        (c) \textbf{Global Interaction Head} (\secref{section:global}):  globally, interactions encode the contexts between every graph components -- nodes and edges, allowing the model to encode richer contextual information. 
        (d) \textbf{Attract \& Repel Loss} (\secref{section:loss}): embeddings of each predicate categories are gathered into compact and well separated clusters by the loss.
        Combining all together, we build an end-to-end, unified framework that predicts a visually grounded scene graph.
    }
    \label{fig:figure_overview}
\end{figure*}

\section{Local-to-Global Interaction Network}
    \label{section:login}
    Based on the analysis in~\secref{section:challenges}, we design a novel framework LOGIN that aims to handle said issues in a bottom-up manner. Each building block in LOGIN is specialized in tackling specific challenges and also works complementary to each other.
    An overview of LOGIN is shown in~\figref{fig:figure_overview}.
    
    \paragraph{Problem Setup}
        Given an image $I$, the detector predicts a set of entity proposals.
        For each entity proposal, it outputs a Region of Interest (RoI) Aligned~\cite{he2017mask} visual appearance feature ${a}_{i} \in \mathbb{R}^{256 \times 7 \times 7}$, a bounding box prediction ${b}_{i} \in \mathbb{R}^4$, and initial classification logit ${c}_{i} \in \mathbb{R}^{151}$.
        In practice, a standard entity detector Faster R-CNN~\cite{ren2015faster} is used as a bounding box model.
        
        Starting from a set of entity proposals (equivalent to a set of nodes in scene graphs), visual features are pooled from the subject and object boxes that form a relationship and from a union box to utilize contextual information (\eg, background) via RoI-Align operation, then predict the node and edge labels through scene graph generation head in turn.
        
        The initial scene graph comprises a set of node representations $\mathcal{N}$ and a set of edge representations $\mathcal{E}$. The ${i}$th initial node representation ${x}_{i} \in \mathcal{N}$ is obtained by fusing three important cues in the image: ${x}_{i} = \phi([{a}_{i} \ || \ b_{i} \ || \ {c}_{i}])$, where $\phi$ is an embedding function and $ || $ denotes concatenation operation. The edge representations $\mathcal{E}$ are obtained through several stages of process that will be described in the following.
        
        The final scene graph is composed of a set of node label distributions $\mathfrak{N} \in \mathbb{R}^{N \times 151}$ (including \textit{no-object} class) and a set of edge label distributions $\mathfrak{E} \in \mathbb{R}^{M \times 51}$ (including \textit{no-relation} class), where $N$ and $M$ is the number of total nodes and edges respectively.
    
    \subsection{Local Interaction Head}
    \label{section:local}
        \begin{figure}[t!]
            \centering
            \includegraphics[width=\linewidth]{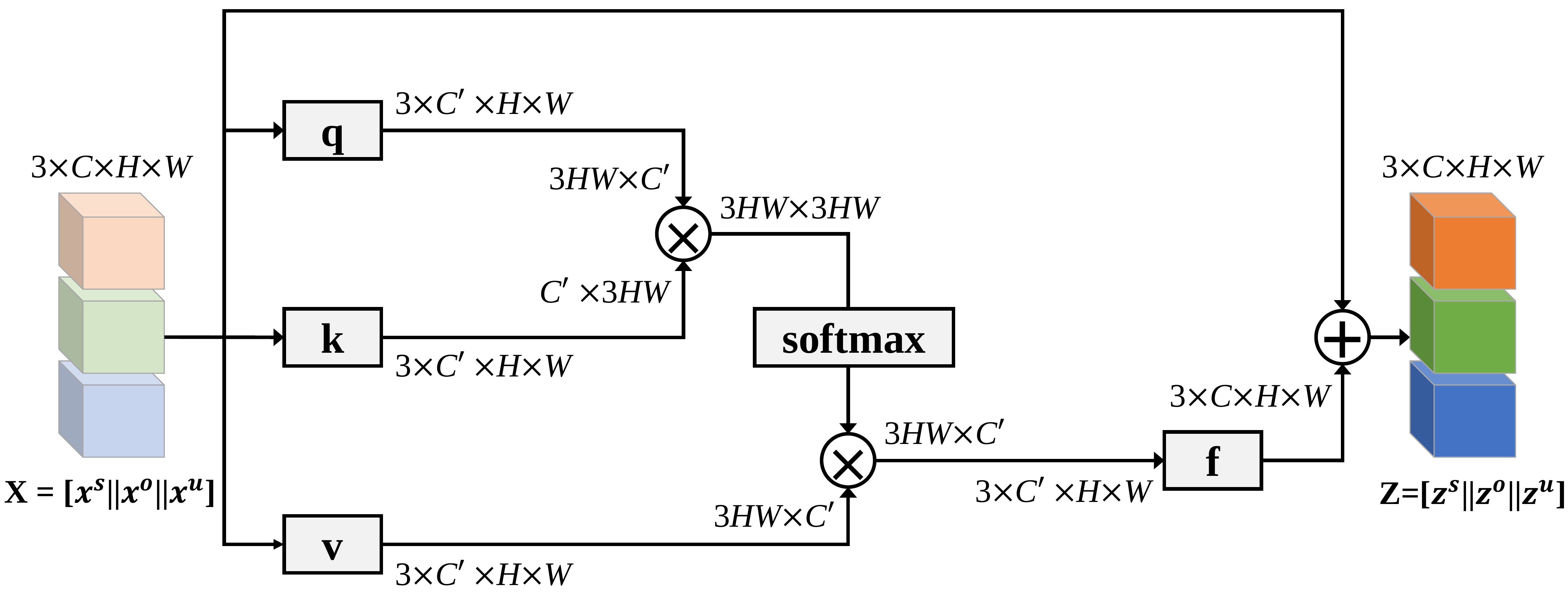}
            \captionsetup{font=footnotesize}
            \caption{
                \textbf{Local-Interaction Head learns what (3$\times$C) and where (H$\times$W) to attend.} It adaptively learns to emphasize informative representation between three entity-level instances by weighing how much each pair-wise interaction contributes to relational representation.
            }
            \label{fig:local_interact}
        \end{figure}
        
        We posit the underlying vulnerability to ambiguity stems from the inability to capture subtle yet discriminative representations.
        Inspired by the recent successes of attention based fine-grained recognition works~\cite{fu2017look, zheng2017learning, wei2018mask, wang2018non, woo2018cbam, zheng2019looking, wang2020axial}, where the intra-class variance is usually high and vice versa for inter-class, we adopt the idea of attention mechanism.
        In particular, as for Local-Interaction Head (LIH), we formulate the instance-level interaction as a non-local operation~\cite{wang2018non}. LIH learns to highlight relationship-centric representations and suppress the noise since the non-local operation considers all individuals to compute responses to the target individual.
        
        The intensity of pair-wise interaction is calculated over the three entity (node) features $\{x^s$, $x^o$, $x^u\}$ (each refers to subject, object, and union) (see~\figref{fig:local_interact}). Given concatenated features $\textbf{X} = [{x}^{s} || {x}^{o} || {x}^{u}]$, LIH outputs refined features $\textbf{Z} = [{z}^{s} || {z}^{o} || {z}^{u}]$.
        The interaction intensity between $i$ and $j$th individual is computed by the embedded gaussian kernel ($e^{\textbf{q}(\cdot)}$, $e^{\textbf{k}(\cdot)}$) and normalized by the sum:
        \begin{eqnarray}
            \begin{aligned}
                & \alpha_{ij} = {{e^{\textbf{q}(x_i)^T \textbf{k}(x_j)}} \over {\sum_{\forall j}} e^{\textbf{q}(x_i)^T \textbf{k}(x_j)}}\,.
            \end{aligned}
            \label{eq:3}
        \end{eqnarray}
        The interaction intensity $\alpha_{ij}$ is multiplied with the representation of $i$th individual $\textbf{v}(x_i)$ followed by a transformation function $\textbf{f}(\cdot)$. The output of LIH opeartion is given by:
        \begin{eqnarray}
            \begin{aligned}
                & z_{ij} = \textbf{f} (\alpha_{ij}^{T} \textbf{v}(x_i)) + x_i \,.
            \end{aligned}
            \label{eq:4}
        \end{eqnarray}
        For the sake of better gradient flow while learning the LIH, a residual-connection ($+ x_i$) is added.
        In practice, $1 \times 1 \times 1$ convolutional operations is used for all embedding functions ($\textbf{q}(\cdot)$, $\textbf{k}(\cdot)$, $\textbf{v}(\cdot)$, $\textbf{f}(\cdot)$). 
        
    \subsection{Encoding Direction-Awareness}
    \label{section:direction}
        \begin{figure}[t!]
            \centering
            \includegraphics[width=\linewidth]{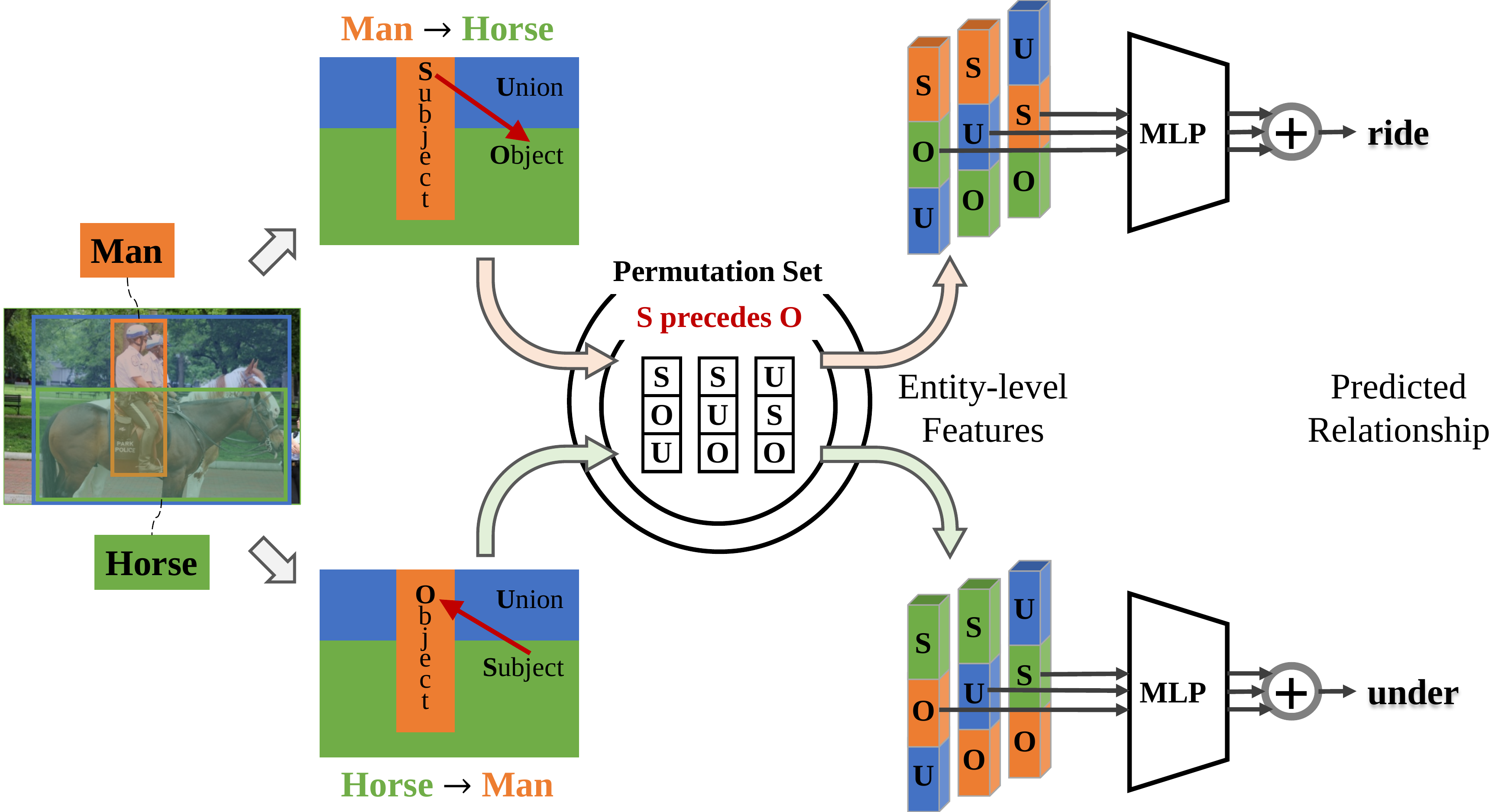}
            \captionsetup{font=footnotesize}
            \caption{
                \textbf{Direction-sensitive encoding.} Between a pair of entities, the subject and the object can be switched, and two opposite-direction relationships are usually asymmetric.
                For example, the relationships of \texttt{Man} $\to$ \texttt{Horse} and \texttt{Horse} $\to$ \texttt{Man} are generally different. 
                For both-side relationships, we fed all possible \textit{permutations} of entity-level features (\eg, subject, object and union) that satisfy the conditions of ``subject precedes object" into the same MLP.
                Note that although both relationships follow the same condition, color combinations (\eg, orange, green, blue) vary with direction.
                The final relationship is predicted after summing all the outputs of the MLP.
                During training, the MLP learns to generate different outputs for the opposite-direction relationships, thus it becomes aware of the relational direction.
            }
            \label{fig:direction}
        \end{figure}
        Before moving on to the next stage, there is an open choice on how to fuse three instance features $\{{z}^{s}, {z}^{o}, {z}^{u}\}$ obtained earlier to initialize a graph-level predicate (edge) representation.
        As suggested in~\cite{santoro2017simple}, summing up all possible permutations of instance features could be a generic method for relational inference.
        The effectiveness of using permutation has been empirically demonstrated in prior works~\cite{zaheer2017deep, zhang2019deep, lee2019set}, but the directionality cannot be guaranteed since summing is commutative. In other words, the permutation sets are identical even if the ordering of two instances are reversed (\ie, the identity of subject and object are switched).
        A simple sidestep is to use the embeddings of concatenated instance features, which is the form used in the several previous works~\cite{zhang2017visual, yang2018shuffle, zellers2018neural, zhang2019graphical, tang2019learning}. However, this also has the disadvantage of losing the benefits of permutation.
        
        We would like the LOGIN to be \textit{equivariant} under the subject-object ordering (\ie, relational direction) while being \textit{invariant} to the permutation.
        Let the interaction between two entities as a set of permutations, $S$, and directional relationship as any subset except the empty and the universal set, ${r}_{i} \subset {S}, \ where \ {r}_{i} \neq \phi \ and \ {r}_{i} \neq {S}, \ i \in \{f, b\}$. If the two opposite-sided relationship (forward and backward) subsets are disjoint, ${r}_{f} \cap {r}_{b} = \phi$, and their union is universal, ${r}_{f} \cup {r}_{b} = {S}$, the relationship encoding of two subsets can always be semantically distinguished.
        Under these premises, we specifically use a regulated set of permutations in which the subject always precedes the object -- $\{SOU, SUO, USO\}$ -- to represent a relationship in one direction. This simple strategy guarantees that subject only appears in the first two bins, and that object only appears in the last two bins. Thus the model can clearly distinguish between forward and backward relationships while sharing entity instance features.
        Note that this strategy is just a straightforward method to make half of the entire permutation set represent the forward direction and the other half represent the backward direction, and it does not matter which combination of permutations is used.
        
        Formally, the three instances from an input set $\{z^s$, $z^o$, $z^u\}$ are concatenated in a constrained order, providing inherent bias of the directionality. They are then transformed via shared MLP (denoted as $\varphi$ in the below equation) and additively fused to make the predicate representation invariant to the input permutations. $i$ th predicate representation ${e}_{i} \in \mathcal{E}$ can be obtained as:
        \begin{eqnarray}
            \begin{aligned}
            	  {e}_{i} = \sum_{j,k,l \in \{z^s, z^o, z^u\}} \varphi([j \ || \ k \ || \ l]),&\\
            	  \text{where $z^s$ precedes $z^o$ and $j\neq k \neq l$.}&
            \end{aligned}
            \label{eq:5}
        \end{eqnarray}
    
    \subsection{Global Interaction Head}
    \label{section:global}
        \begin{figure}[t!]
            \centering
            \includegraphics[width=\linewidth]{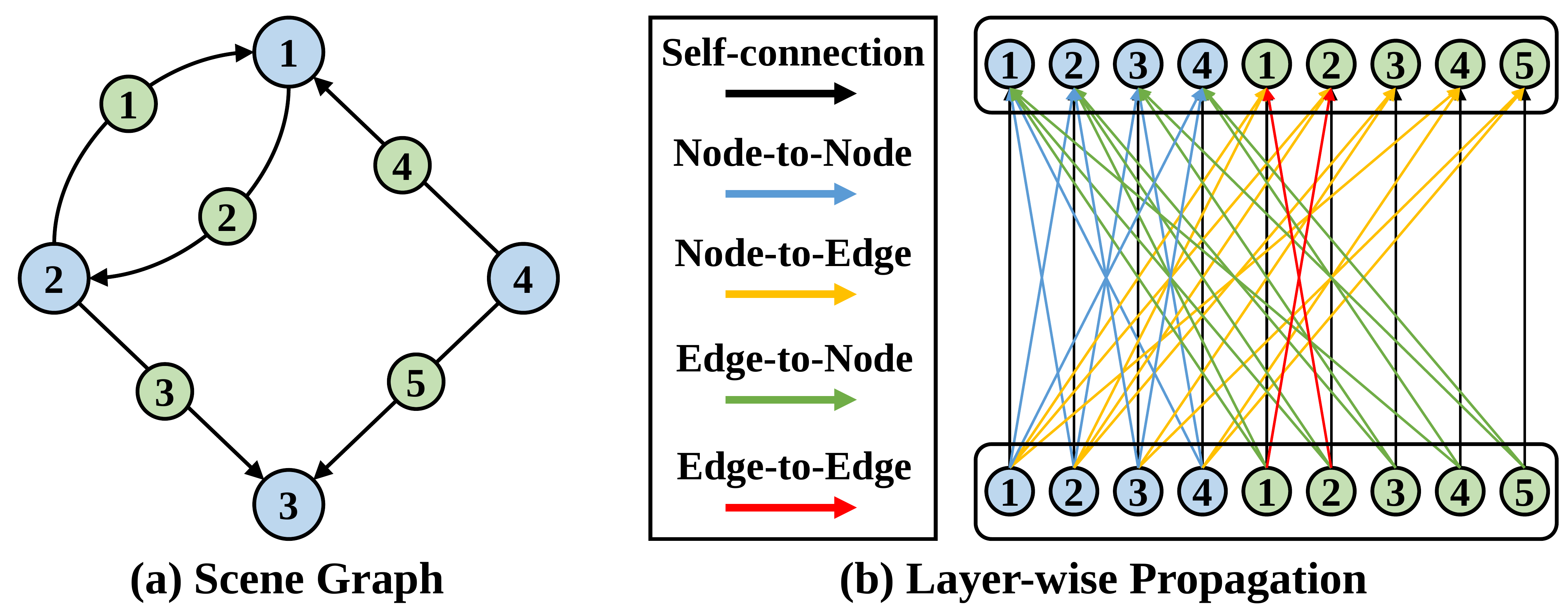}
            \captionsetup{font=footnotesize}
            \caption{
                \textbf{An illustration of layer-wise propagation of Global-Interaction Head.} Nodes and edges are colored in blue and green respectively. The layer-wise context propagation of GIH in scene graph (a) can be represented as a bipartite graph in (b). As a comparison, conventional GCN~\cite{kipf2017semi} and GAT~\cite{velivckovic2018graph} only consider node-wise propagation (black and blue edges) and are unable to leverage edge information. An experimental comparison of GIH with GCN and GAT is in~\tabref{tab:gih}.
            }
            \label{fig:global_interact}
        \end{figure}
        
        \begin{figure*}[t!]
            \centering
            \includegraphics[width=\linewidth]{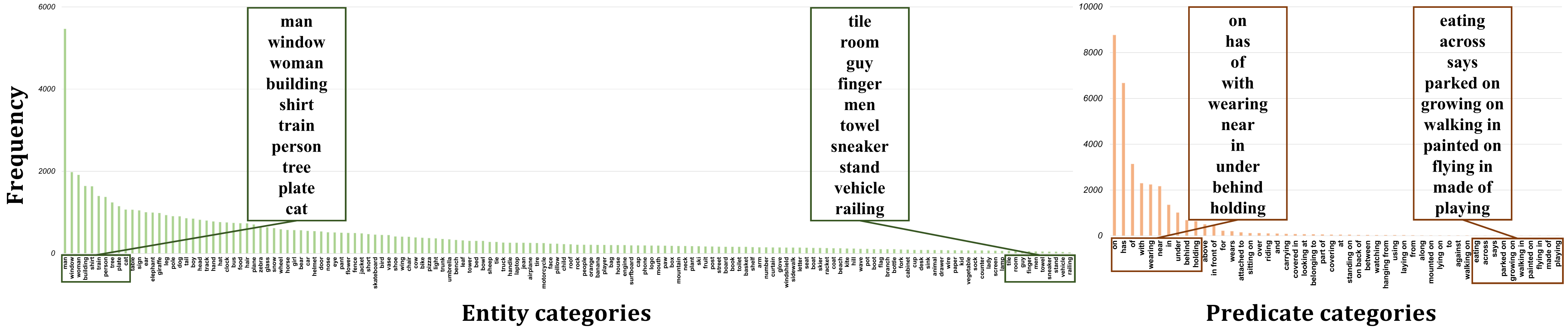}
            \captionsetup{font=footnotesize}
            \caption{
                \textbf{The distribution of categories in the BR Dataset.} (a) Frequency of entity categories and (b) predicate categories. For both entity and predicate, the top-10 categories and the bottom-10 categories are highlighted based on frequency. Axis labels best viewed zoomed in on screen.
            }
            \label{fig:bidirectional_dist}
        \end{figure*}
        
        From a graph perspective, the fully-connected layer can be seen as the most basic form of message passing network with all nodes connected, but it is known to be not effective in learning the graph. For effective context aggregation, well structuring the message paths (\ie, connectivity between nodes) is the key issue. Based on the observation in \secref{section:contexts}, we design a Global Interaction Head (GIH) that enables effective message flow between informative graph components.
        We formulate the graph-level interaction with global message passing scheme~\cite{kipf2017semi, gilmer2017neural, velivckovic2018graph}. 
        
        To maintain a structured representation of a scene graph, we utilize local connectivity information in the form of a block matrix with four quadrants $A \in \mathbb{R}^{(N+M)\times (N+M)}$. Each quadrant from top left to bottom right indicates whether the \textit{node}-\textit{node} ($A_{n-n} \in \mathbb{R}^{N \times N}$), \textit{node}-\textit{edge} ($A_{n-e} \in \mathbb{R}^{N \times M}$), \textit{edge}-\textit{node} ($A_{e-n} \in \mathbb{R}^{M \times N}$), and \textit{edge}-\textit{edge} ($A_{e-e} \in \mathbb{R}^{M \times M}$) are connected (1) or not (0) -- the number of nodes and edges are denoted as $N$ and $M$ respectively. We consider that all node pairs and node-edge pairs that make up a relationship are interconnected. In the case of \textit{edge}-\textit{edge}, they are considered to be connected when the opposite direction edge exists, although there is no explicit connection on the graph (see~\figref{fig:global_interact}).
        \begin{eqnarray}
            \begin{aligned}
        &A = \left[\begin{array}{r@{}c|c@{}l}
                      & {A}_{n-n} \rule[-1ex]{0pt}{2ex}
                      & {A}_{n-e} & \rlap{\kern10mm}\\\hline
                      & {A}_{e-n}
                      & {A}_{e-e} &  \rlap{\kern10mm}
                    \end{array} \right] \,.
            \end{aligned}
            \label{eq:6}
        \end{eqnarray}
        To preserve the original message, identity matrix (self-connection) is added to $A$, resulting $\tilde{A} = A + I$. An initial graph-level feature matrix $\mathcal{G}^{(0)} \in \mathbb{R}^{(N+M) \times D}$ is defined as:
        \begin{eqnarray}
            \begin{aligned}
                \mathcal{G}^{(0)} = \left[\begin{array}{c}
                                      \mathcal{N} \\\hline
                                      \mathcal{E}
                                    \end{array} \right] \,.
            \end{aligned}
            \label{eq:7}
        \end{eqnarray}
        The $l$th layer-wise propagation rule for GIH is defined as:
        \begin{equation}
            \resizebox{.91\linewidth}{!}{$
                \displaystyle
                \mathcal{G}^{(l)} =\left\{
                    \begin{array}{ll}
                        \text{max} \left(0, \tilde{A}\mathcal{G}^{(l-1)}W^{(l-1)} \right), \text{$l$ = odd.} \\
                        \mathcal{G}^{(l-2)}+\text{max} \left(0, \tilde{A}\mathcal{G}^{(l-1)}W^{(l-1)} \right), \text{$l$ = even.} 
                    \end{array}\right.\\
            $}
            \label{eq:8}
        \end{equation}%
        
        We add residual connections between the layers for a better optimization~\cite{li2019deepgcns}. Multi-layer GIH can perform long-range multi-hop communication, effectively modeling the desired higher-order relational reasoning. While training, weight matrix $W \in \mathbb{R}^{D \times D}$ is learned by gradient.
        
        Finally, the upper $N$ rows ($\mathcal{N}^{'} \in \mathbb{R}^{N \times D}$) and the lower $M$ rows ($\mathcal{E}^{'} \in \mathbb{R}^{M \times D}$) of the output matrix are softmax-ed and used to predict entity and predicate labels.
    
    \subsection{Attract \& Repel Loss}
    \label{section:loss}
        We introduce an Attract \& Repel Loss to explicitly handle the intra- and inter-class variance.
        The conceptual mechanism of Attract \& Repel loss is shown in~\figref{fig:attract_repel_loss}.
        In a nutshell, if the identities of the input and reference embeddings are the same (\ie, category matches), the loss forces them to \emph{attract} each other; otherwise, the loss compels them to \emph{repel} each other.
        The reference embeddings can be divided into two types: we refer to the running mean of the matched reference embeddings as positive ($pos$), and negative ($neg$) for that of non-matched.
        Note that the reference type is only an abstract distinction and can be vary depending on the identity of the input embedding.
        As the input embeddings are learned to approach the positive and move farther away from the negatives, the distribution within categories becomes dense, and between categories becomes sparse.
        As a result, the loss gathers the embeddings of each class into compact and well-separated clusters.
        Since most bidirectional relationships are asymmetric (\ie, identities of predicates in opposite directions are mostly different) as we have seen in~\secref{section:asymmetry}, the loss has the potential benefit in predicting predicates in opposite directions differently.
        Formally, at $t$th batch, given a set of predicate embeddings $\mathcal{E}^{'(t)} = \{{e}_{1}^{'(t)}, ... , {e}_{M}^{'(t)}\}$, a set of references $\mathcal{R}^{(t)} = \{r_{1}^{(t)}, ... , r_{51}^{(t)}\}$ is adjusted with the following update rule:
        \begin{eqnarray}
            \begin{aligned}
            	  & {r}_{m}^{(t)} = {{r}_{m}^{(t-1)} * \mathbb{N}({r}_{m}^{(t-1)}) + \sum_{i \in pos} {e}_{i}^{'(t)} - \sum_{j \in neg} {e}_{j}^{'(t)} \over {\mathbb{N}({r}_{m}^{(t-1)}) + \mathbb{N}(pos) + \mathbb{N}(neg)}} \,,\\
            \end{aligned}
            \label{eq:9}
        \end{eqnarray}
        where $\mathbb{N}(\cdot)$ denotes the number of embeddings considered in reference update. Finally, the input predicate embeddings are adjusted with the following Attract \& Repel loss:
        \begin{eqnarray}
            \begin{aligned}
                \mathcal{L}_{ar} = & \sum_{m} \sum_{i \in pos} \left({1 - {{{r}_{m}^{(t)} \cdot {e}_{i}^{'(t)}} \over {|{r}_{m}^{(t)}| \ |{e}_{i}^{'(t)}|}}} \right) + \\
                & \sum_{m} \sum_{j \in neg} \left( {{{r}_{m}^{(t)} \cdot {e}_{j}^{'(t)}} \over {|{r}_{m}^{(t)}| \ |{e}_{j}^{'(t)}|}} \right) \,.
            \end{aligned}
            \label{eq:10}
        \end{eqnarray}
    
    \subsection{Loss Function}
    \label{section:loss_func}
        LOGIN can be trained in an end-to-end manner, allowing the network to predict bounding boxes, entity categories, and relationship categories at once.
        The total loss function for an image is defined as:
        \begin{eqnarray}
        \mathcal{L}_{image} = \mathcal{L}_{ent} + \mathcal{L}_{pred} + \mathcal{L}_{ar} \,, \label{eq:11}
        \end{eqnarray}
        where $\mathcal{L}_{ent}$ and $\mathcal{L}_{pred}$ are both cross-entropy loss for entity and predicate classification, respectively. $\mathcal{L}_{ar}$ stands for the Attract \& Repel loss. By default, hyperparameters of joint loss function are set as 1:1:1.

%% file: 5.experiments.tex
\section{Experiments}
    \label{section:experiments}
    \begin{figure}[t!]
        \centering
        \includegraphics[width=\linewidth]{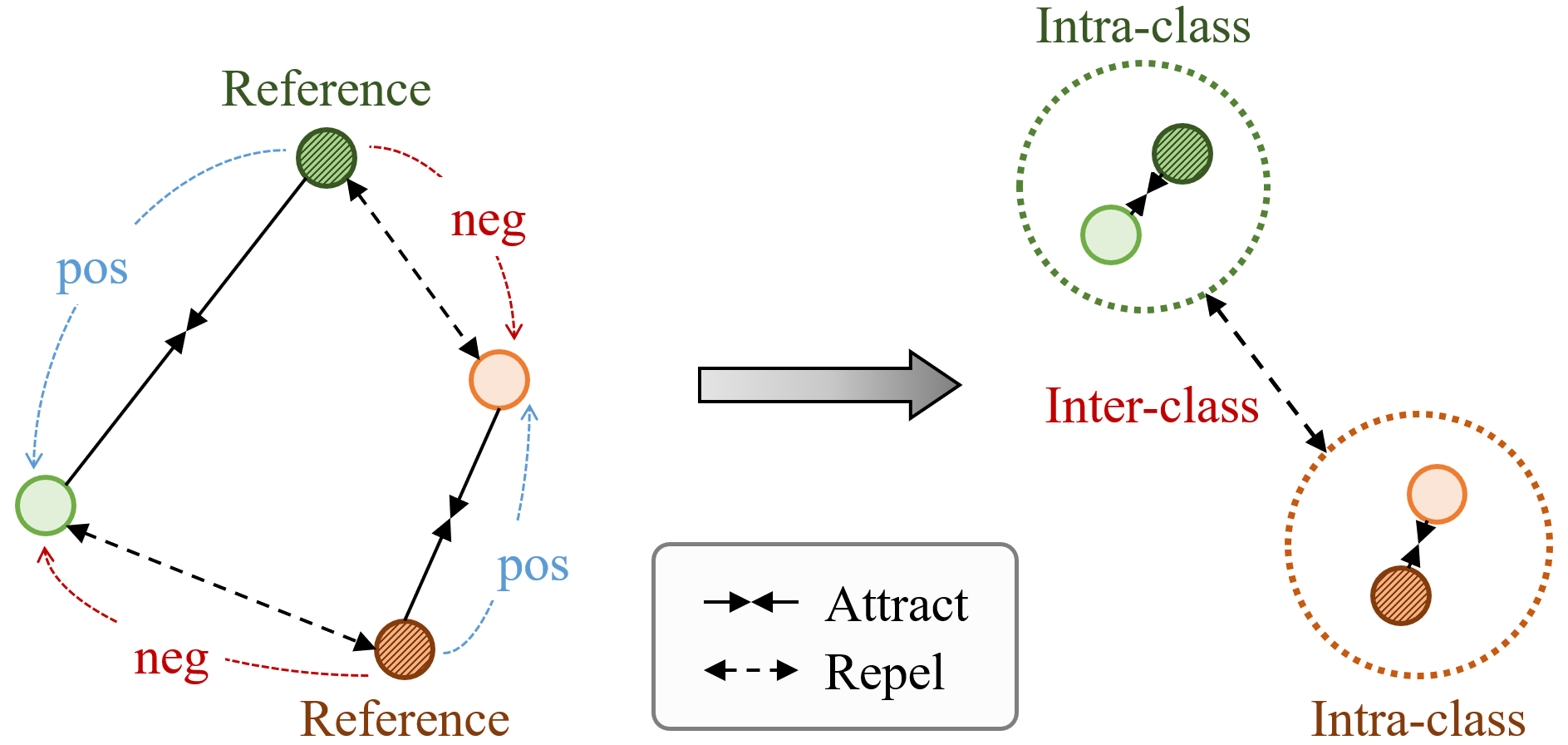}
        \captionsetup{font=footnotesize}
        \caption{
            \textbf{Mechanism of Attract \& Repel loss.} The reference embeddings attract the positives and repel the negatives, making intra-class distribution dense and inter-class distribution sparse. For simplicity, we visualize only one positive and negative instance.
        }
        \label{fig:attract_repel_loss}
    \end{figure}
        
    \begin{table*}[t!]
        \captionsetup{font=footnotesize}
        \caption{
            \textbf{Comparison with the state-of-the arts on Visual Genome benchmark.} \textit{R@k} denotes Recall in the top-$k$ predictions.
        }
        \centering
        \resizebox{0.85\linewidth}{!}{
        \begin{tabular}{@{}cccccccccccc@{}}
            \toprule[0.15em]
            \multicolumn{3}{c}{} & \multicolumn{3}{c}{\textit{SGGen}} & \multicolumn{3}{c}{\textit{SGCls}} & \multicolumn{3}{c}{\textit{PredCls}} \\ \cmidrule(l){4-12} 
            \multicolumn{3}{c}{Models} & R@20 & R@50 & R@100 & R@20 & R@50 & R@100 & R@20 & R@50 & R@100 \\ \midrule
            \multicolumn{3}{c}{\textsc{Imp}~\cite{xu2017scene}} & - & 3.4 & 4.2 & - & 21.7 & 24.4 & - & 44.8 & 53.0 \\
            \multicolumn{3}{c}{\textsc{MotifNet}~\cite{zellers2018neural}} & 21.4 & 27.2 & 30.3 & 32.9 & 35.8 & 36.5 & 58.5 & 65.2 & 67.1 \\
            \multicolumn{3}{c}{\textsc{Graph R-CNN}~\cite{yang2018graph}} & - & 11.4 & 13.7 & - & 29.6 & 31.6 & - & 54.2 & 59.1 \\
            \multicolumn{3}{c}{\textsc{Kern}~\cite{chen2019knowledge}} & - & 27.1 & 29.8 & - & 36.7 & 37.4 & - & 65.8 & 67.6 \\
            \multicolumn{3}{c}{\textsc{Cmat}~\cite{chen2019counterfactual}} & \textit{22.1} & 27.9 & 31.2 & 35.9 & \textbf{39.0} & \textit{39.8} & 60.2 & 66.4 & 68.1 \\
            \multicolumn{3}{c}{\textsc{VCTree}~\cite{tang2019learning}} & 22.0 & 27.9 & 31.3 & 35.2 & 38.1 & 38.8 & 60.1 & 66.4 & 68.1 \\
            \multicolumn{3}{c}{\textsc{RelDN}~\cite{zhang2019graphical}} & 21.1 & \textbf{28.3} & \textbf{32.7} & \textbf{36.1} & 36.8 & 36.8 & \textbf{66.9} & \textbf{68.4} & \textit{68.4} \\
            \midrule
            \rowcolor{Gray}
            \multicolumn{3}{c}{\textsc{\textbf{LOGIN (Ours)}}} & \textbf{22.2} & \textit{28.2} & \textit{31.4} & \textit{35.5} & \textit{38.8} & \textbf{40.5} & \textit{61.1} & \textit{66.6} & \textbf{68.7} \\
            \bottomrule[0.15em]
        \end{tabular}
        }
        \label{tab:recall}
    \end{table*}
    
    \begin{table*}[t!]
        \captionsetup{font=footnotesize}
        \caption{
            \textbf{The SGG results on mean Recall (mR@K).} \textit{mR@k} denotes average \textit{R@K} over all predicate categories.
        }
        \centering
        \resizebox{\linewidth}{!}{
        \begin{tabular}{@{}cccccccccccc@{}}
            \toprule[0.15em]
            \multicolumn{3}{c}{} & \multicolumn{3}{c}{\textit{SGGen}} & \multicolumn{3}{c}{\textit{SGCls}} & \multicolumn{3}{c}{\textit{PredCls}} \\ \cmidrule(l){4-12} 
            \multicolumn{3}{c}{Models} & mR@20 & mR@50 & mR@100 & mR@20 & mR@50 & mR@100 & mR@20 & mR@50 & mR@100 \\ \midrule
            \multicolumn{3}{c}{\textsc{Imp}~\cite{xu2017scene}} & - & 3.8 & 4.8 & - & 5.8 & 6.0 & - & 9.8 & 10.5 \\
            \multicolumn{3}{c}{\textsc{MotifNet}~\cite{zellers2018neural}} & 4.2 & 5.7 & 6.6 & 6.3 & 7.7 & 8.2 & 10.8 & 14.0 & 15.3 \\
            \multicolumn{3}{c}{\textsc{Kern}~\cite{chen2019knowledge}} & - & 6.4 & 7.3 & - & 9.4 & 10.0 & - & 17.7 & 19.2 \\
            \multicolumn{3}{c}{\textsc{VCTree}~\cite{tang2019learning}} & 5.2 & 6.9 & 8.0 & 8.2 & 10.1 & 10.8 & 14.0 & 17.9 & 19.4 \\
            \midrule
            \rowcolor{Gray}
            \multicolumn{3}{c}{\textsc{\textbf{LOGIN (Ours)}}} & \textbf{5.9} & \textbf{7.7} & \textbf{9.1} & \textbf{8.6} & \textbf{11.2} & \textbf{12.4} & \textbf{16.0} & \textbf{19.2} & \textbf{22.3} \\
            \bottomrule[0.15em]
        \end{tabular}
        }
        \label{tab:mean_recall}
    \end{table*}
    
    In this section, we conduct comprehensive studies to validate the efficacy of LOGIN.
    We perform extensive ablation experiments to thoroughly demonstrate the effectiveness of each building block of LOGIN.
    LOGIN is evaluated on Visual Genome~\cite{krishna2017visual} benchmark and achieves state-of-the-art results.
    Notably, in our proposed Bidirectional Relationship Classification (BRC) task, LOGIN successfully distinguishes asymmetric relationships and is more accurate than existing methods.
    
    The model referred to as the \textsc{Baseline} in this section is a model without any proposed design principles. It directly predicts the entity and predicate categories from the RoI-Aligned visual features of entity instances and that of union of two entity instances, respectively.
    
    \subsection{Settings}
    \label{section:setting}
        \paragraph{Model Parameter and Training Details}
            For a fair comparison, most of the settings and details follow pioneer work~\cite{xu2017scene, zellers2018neural}.
            We adopt the Faster R-CNN~\cite{ren2015faster} detector with VGG backbone~\cite{simonyan2014very}.
            Following~\cite{zellers2018neural}, we use per-class NMS to reduce the number of entity proposals. The number of entity proposals is 64 (\ie, $N$=64).
            We optimize the model using SGD with the following details: initial learning rate (1e-3), momentum (0.9), and weight decay (5e-4).
            We first pre-train the detector on Visual Genome Dataset and then train the proposed scene graph generation head while fixing the detector weight.
            To model geometric relationships, we first concatenate two extra channels with coordinates hard-coded (2$\times$7$\times$7) to the initial visual representation and then pass them through a convolutional layer~\cite{liu2018intriguing}. As for the Attract \& Repel loss, we sample negatives by the number of positives to avoid being heavily affected by negatives.
        
        \begin{table}[t!]
            \captionsetup{font=footnotesize}
            \caption{
                \textbf{Comparison with recent approaches in the BRC task.}
            }
            \centering
            \resizebox{\linewidth}{!}{
            \begin{tabular}{@{}cccccc@{}}
                \toprule[0.15em]
                \multicolumn{2}{c}{} & \multicolumn{4}{c}{\textit{PredCls}} \\ \cmidrule(l){3-6} 
                \multicolumn{2}{c}{Models} & pR@2 & pR@4 & pR@8 & pR@16 \\ \midrule
                \multicolumn{2}{c}{\textsc{Imp}~\cite{xu2017scene}} & 6.3 & 9.1 & 12.2 & 15.0 \\
                \multicolumn{2}{c}{\textsc{MotifNet}~\cite{zellers2018neural}} & 7.7 & 11.5 & 15.9 & 19.5 \\
                \multicolumn{2}{c}{\textsc{Graph R-CNN}~\cite{yang2018graph}} & 7.9 & 11.7 & 16.3 & 21.0 \\
                \multicolumn{2}{c}{\textsc{Kern}~\cite{chen2019knowledge}} & 7.7 & 12.1 & 16.7 & 20.7 \\
                \multicolumn{2}{c}{\textsc{VCTree}~\cite{tang2019learning}} & 8.0 & 11.9 & 16.1 & 21.0 \\
                \multicolumn{2}{c}{\textsc{RelDN}~\cite{zhang2019graphical}} & 8.0 & 12.5 & 16.4 & 20.8 \\
                \midrule
                \rowcolor{Gray}
                \multicolumn{2}{c}{\textsc{\textbf{LOGIN (Ours)}}} & \textbf{8.6} & \textbf{13.1} & \textbf{17.6} & \textbf{21.1} \\
                \bottomrule[0.15em]
            \end{tabular}
            }
            \label{tab:brc}
        \end{table}
        
        \paragraph{VG Dataset} 
            We train and evaluate LOGIN on Visual Genome (VG) Dataset~\cite{krishna2017visual}. We use the publicly released pre-processed data (train and test split is 75K and 32K)~\cite{xu2017scene}. The number of entity and predicate categories are 150 and 50, respectively.
            
        \paragraph{BR Dataset}
            We build a Bidirectional Relationship (BR) dataset to evaluate the direction awareness of the model.
            The BR dataset is a subset of VG dataset and is created by filtering out relationships with only one edge between the two nodes.
            As a result, the BR dataset always includes the relationships that have two bidirectional edges between the nodes (\eg, $\texttt{man} \xrightarrow{riding} \texttt{horse}$, $\texttt{horse} \xrightarrow{under} \texttt{man}$).
            As shown in~\figref{fig:Motivation}, about 93\% of bidirectional edges form different relationships depending on the direction (\ie, direction-sensitive), and only about 7\% of bidirectional edges have the same relationship regardless of direction (\ie, direction-agnostic).
            The distribution of BR Dataset is shown in \figref{fig:bidirectional_dist}. Here, the five most frequent entity categories and predicate categories are ``\texttt{man} (5466), \texttt{window} (1976), \texttt{woman} (1912), \texttt{building} (1640), \texttt{shirt} (1632)'', and ``\texttt{on} (8766), \texttt{has} (6669), \texttt{of} (3137), \texttt{with} (2292), \texttt{wearing} (2238)'', respectively. Note that the top-5 predicate categories account for about 73\% of the total predicates.
            This shows that the dominant predicate categories are often used in various contexts repeatedly, implying that variance may be high even within the same predicate category.
            That is to say, this biasness supports our argument that dealing with \textit{ambiguity} issue is essential.
            
        \begin{table*}[t!]
            \captionsetup{font=footnotesize}
            \caption{\textbf{(a) Ablation studies on network design. (b) Optimal variable search.}}
            \centering
            \begin{minipage}[t]{0.66\linewidth}
                \centering
                \begin{tabular}{@{}c@{}}
                    \textrm{(a) Ablation Studies}
                \end{tabular}
                \resizebox{\linewidth}{!}{
                \begin{tabular}{@{}cccccccccccc@{}}
                    \toprule[0.15em]
                    & \multicolumn{4}{c}{Ablations} & \multicolumn{3}{c}{\textit{SGCls}} & \multicolumn{4}{c}{\textit{PredCls}} \\ \cmidrule(l){6-12}
                    Exp &  LIH     &  DSE     &  GIH     & $\mathcal{L}_{ar}$ & R@20  & R@50  & R@100 & pR@2  & pR@4  & pR@8  & pR@16 \\ \midrule
                    1   &          &          &          &          & 30.8  & 34.7  & 36.2  &  0.2  &  0.4  &  0.7  &  1.3 \\
                    2   &\checkmark&          &          &          & 33.5  & 37.5  & 39.6  &  8.0  &  11.5 & 17.0  & 20.3 \\
                    3   &          &\checkmark&          &          & 31.8  & 36.2  & 37.7  &  8.1  &  11.5 & 16.4  & 20.1 \\
                    4   &          &          &\checkmark&          & 33.2  & 38.4  & 39.9  &  8.2  &  12.0 & 16.9  & 20.6 \\
                    5   &          &          &          &\checkmark& 31.2  & 35.6  & 36.9  &  7.4  &  11.1 & 16.0  & 19.7 \\
                    6   &\checkmark&\checkmark&\checkmark&          & 34.4  & 38.5  & 40.3  &  8.4  &  12.9 & 17.5  & 21.0 \\
                    \rowcolor{Gray}
                    7   &\checkmark&\checkmark&\checkmark&\checkmark& \textbf{34.5} & \textbf{38.8} & \textbf{40.5} & \textbf{8.6} & \textbf{13.1} & \textbf{17.6} & \textbf{21.1} \\
                    \bottomrule[0.15em]
                \end{tabular}
                }
            \end{minipage}\hfill%
            \begin{minipage}[t]{0.32\linewidth}
                \centering
                \begin{tabular}{@{}c@{}}
                    \textrm{(b) Optimal Variable Search}
                \end{tabular}
                \resizebox{\linewidth}{!}{
                \begin{tabular}{@{}ccccc@{}}
                    \toprule[0.15em]
                    \multicolumn{2}{c}{} & \multicolumn{3}{c}{\textit{SGCls}} \\ \cmidrule(l){3-5}
                    \multicolumn{2}{c}{Variables} & R@20 & R@50 & R@100 \\ \midrule
                    \multicolumn{1}{c|}{\multirow{3}{*}{\rot{Feature}}} & \textsc{avgpool} & 34.3  & 38.4  & 40.2 \\
                    \multicolumn{1}{c|}{} & \textsc{maxpool} & 34.1  & 38.4  & 40.1 \\
                    \multicolumn{1}{c|}{} & \cellcolor{Gray} \textsc{\textbf{flatten}} & \cellcolor{Gray} \textbf{34.5}  & \cellcolor{Gray} \textbf{38.8}  & \cellcolor{Gray} \textbf{40.5}  \\ \midrule
                    \multicolumn{1}{c|}{\multirow{3}{*}{\rot{GIH}}} & \textsc{2-layers} & 34.1 & 38.4 & 39.9 \\
                    \multicolumn{1}{c|}{} & \cellcolor{Gray} \textsc{\textbf{4-layers}} & \cellcolor{Gray} \textbf{34.5} & \cellcolor{Gray} \textbf{38.8} & \cellcolor{Gray} 40.5 \\
                    \multicolumn{1}{c|}{} & \textsc{6-layers} & 34.1 & 38.5 & \textbf{40.6} \\
                    \bottomrule[0.15em]
                \end{tabular}
                }
            \end{minipage}
            \label{tab:ablation}
        \end{table*}
        
        \paragraph{Evaluation Setup}
            Model is evaluated with the following three standard evaluation criteria~\cite{xu2017scene}:
            \begin{enumerate}
            \item Predicate Classification (\textit{PredCls}): Given ground truth boxes and labels, predict edge labels.
            \item Scene Graph Classification (\textit{SGCls}): Given ground truth boxes, predict box and edge labels.
            \item Scene Graph Generation (\textit{SGGen}): Predict boxes, box labels, and edge labels.
            \end{enumerate}
            
            As for SGG, following the prior works~\cite{xu2017scene, zellers2018neural, yang2018graph}, we use Recall@K (R@K) as an evaluation metric since mAP-like metrics are not appropriate due to the sparse annotation in Visual Genome. Specifically, we use image-wise Recall@\{20,50,100\}, which computes the fraction of ground-truth triplets found in the top-$K$ predicted triplets.
            We also adopt the mean Recall@K (mR@K) metric~\cite{tang2019learning, chen2019knowledge, tang2020unbiased} for evaluation, which retrieves each individual predicate and then averages R@K over all predicate categories.
            
            As for BRC, conventional triplet recall-based metrics only consider uni-direction, making it difficult to make a rigorous evaluation of direction awareness. To this end, we have come to introduce a new metric called \textit{pair-wise Recall (pR@K)} that fits the BRC task. The proposed metric is considered to be ``matched" only when bidirectional relationships are both correct. Formally, the pR@K calculates the fraction of the total amount of matched bidirectional relationships (BRs):
            \begin{eqnarray}
                \begin{aligned}
                    \text{pR@K} = {{|\{\text{top-K predicted BRs}\} \cap \{\text{total BRs}\}|} \over {|\{\text{total BRs}\}|}} \,.
                \end{aligned}
                \label{eq:12}
            \end{eqnarray}
            This constraint severely penalizes if the relationship predictions in the opposite direction are the same. Therefore, models without direction awareness cannot receive a high score on this metric.
            For example, if only union features are used, there is no chance that asymmetric relationships are correctly predicted since only the same results are output for BRs.
            To get a high score from this metric, the model needs direction awareness that is essential to correctly predict asymmetric relationships that account for most BRs in the BR datasets.
            Specifically, we use $pR@\{2,4,8,16\}$ in the BRC task since only a few bidirectional relationships are annotated per image ($\sim$3 BRs / image).
            
        \begin{table}[t!]
            \captionsetup{font=footnotesize}
            \caption{
                \textbf{Comparison of feature fusion methods.} (a) Scene Graph Classification results on Visual Genome dataset. (b) Bidirectional Relationship Classification results on BR dataset.
            }
            \centering
            \begin{tabular}{@{}c@{}}
                \textrm{(a) Scene Graph Classification}
            \end{tabular}
            \begin{tabular}{@{}ccccc@{}}
                \toprule[0.15em]
                \multicolumn{2}{r}{} & \multicolumn{3}{c}{\textit{SGCls}} \\ \cmidrule(l){3-5}
                \multicolumn{2}{c}{Fusion} & R@20 & R@50 & R@100 \\ \midrule
                \multicolumn{2}{c}{\textsc{Baseline}} & 30.8 & 34.7 & 36.2 \\ 
                \multicolumn{2}{c}{w/o Permutation} & 33.6 & 38.0 & 39.9 \\
                \multicolumn{2}{c}{Sequential} & 34.1 & 38.4 & 40.1 \\
                \rowcolor{Gray}
                \multicolumn{2}{c}{\textbf{Parallel (Ours)}} & \textbf{34.5} & \textbf{38.8} & \textbf{40.5} \\
                \bottomrule[0.15em]
            \end{tabular}
            \begin{tabular}{@{}c@{}}
                \\
                \textrm{(b) Bidirectional Relationship Classification}
            \end{tabular}
            \begin{tabular}{@{}cccccc@{}}
                \toprule[0.15em]
                \multicolumn{2}{r}{} & \multicolumn{4}{c}{\textit{PredCls}} \\ \cmidrule(l){3-6}
                \multicolumn{2}{c}{Fusion} & pR@2 & pR@4 & pR@8 & pR@16 \\ \midrule
                \multicolumn{2}{c}{\textsc{Baseline}} & 0.2 & 0.4 & 0.7 & 1.3 \\ 
                \multicolumn{2}{c}{w/o Permutation} & 8.2 & 12.2 & 17.0 & 20.6 \\
                \multicolumn{2}{c}{Sequential} & 8.3 & 12.7 & 17.3 & 20.9 \\
                \rowcolor{Gray}
                \multicolumn{2}{c}{\textbf{Parallel (Ours)}} & \textbf{8.6} & \textbf{13.1} & \textbf{17.6} & \textbf{21.1} \\
                \bottomrule[0.15em]
            \end{tabular}
            \label{tab:fusion}
        \end{table}
    
        \begin{figure}[t!]
            \centering
            \includegraphics[width=\linewidth]{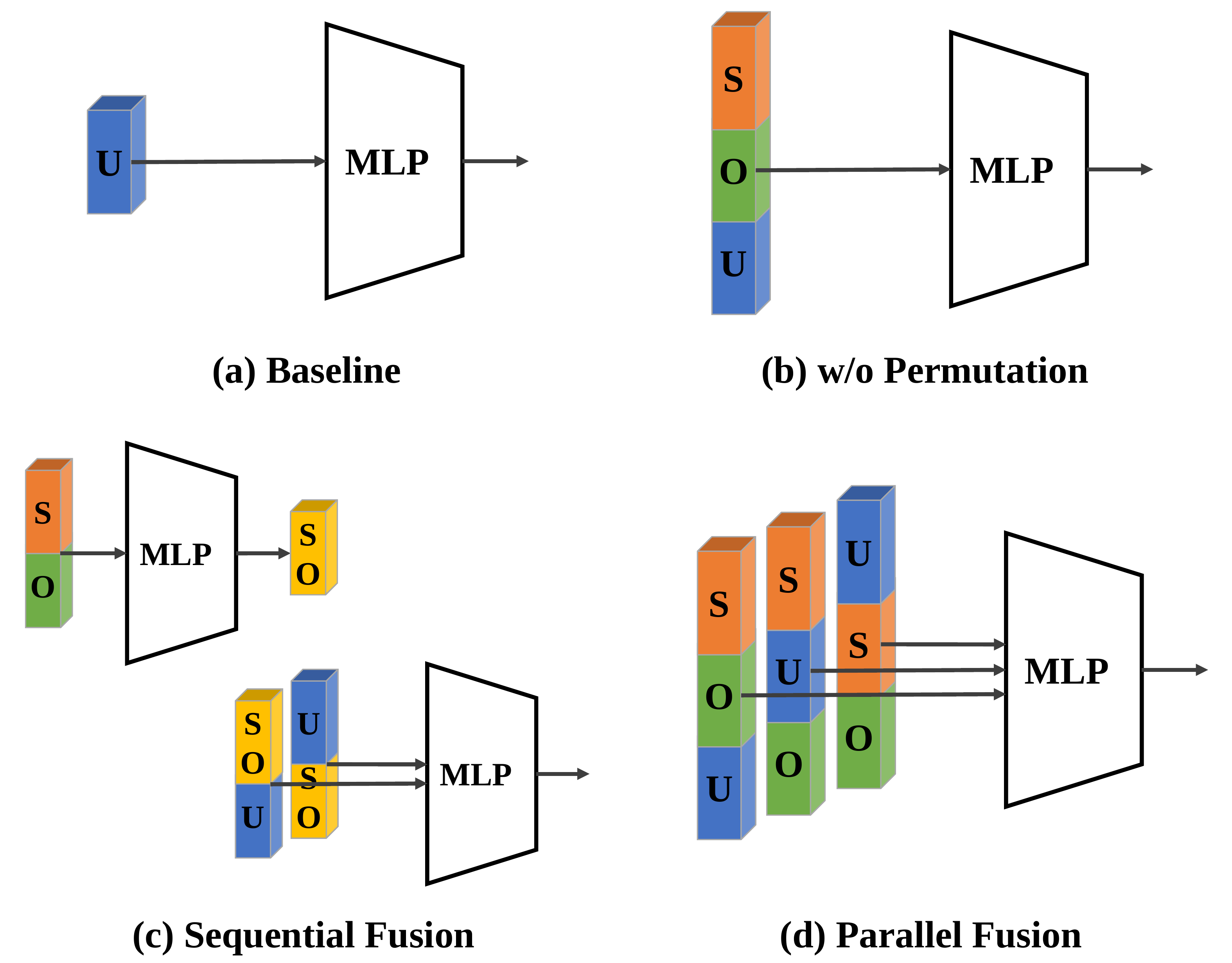}
            \captionsetup{font=footnotesize}
            \caption{
              \textbf{Illustration of feature fusion methods to obtain initial predicate representation.} (a) \textsc{Baseline}: use union feature only. (b) w/o Permutation: concatenate all and fuse them without permutation. (c) Sequential: fuse subject and object first, and then with union. (d) Parallel: fuse all the permutations at once where the subject precedes object. Here, $S$, $O$, $U$ respectively denotes subject, object, and union.
            }
            \label{fig:fusion}
        \end{figure}
    
    \subsection{Comparison with State-of-the-Art}
    \label{section:results}
        \paragraph{Scene Graph Generation (SGG)}
            The Recall performance of the proposed method and existing methods are compared in~\tabref{tab:recall} for each evaluation criterion.
            We compare LOGIN with the recent approaches~\cite{xu2017scene, zellers2018neural, yang2018graph, chen2019knowledge, chen2019counterfactual, tang2019learning, zhang2019graphical}.
            While LOGIN appears to show competitive results against the state-of-the-arts in all criteria, note that there is no specific method that achieves the best performances in every evaluation criteria, making it difficult to judge the superiority among the SGG methods.
            
            We also benchmark LOGIN under the mean Recall (mR@K) criteria. The results are shown in~\tabref{tab:mean_recall}. The mean Recall is measured by averaging the Recall per each class for the entire classes. Therefore, unlike conventional Recall (R@K), it is irrelevant to the number of samples in each class, and even if high performance is obtained in a class with a large number of samples, it is difficult to achieve good values if low performance is obtained in a class with a small number of samples. That is, every class should obtain a good overall recall to achieve high performance. In short, it is important to accurately predict the class with few data, especially in tail, among long-tailed VG dataset. The long-tailed distribution of VG dataset also implies that the dominant predicates frequently appear in multiple contexts. Thus, it is also related to the ambiguity issue.
            We see that LOGIN consistently outperforms recent methods in mean Recall criteria (see~\tabref{tab:top20} for Recall of individual predicate), implying that our system effectively deals with ambiguity issue.
        
        \paragraph{Bidirectional Relationship Classification (BRC)}
            To independently evaluate the direction-awareness of the model, we specifically use \textit{PredCls} criteria, which is orthogonal to the entity detection.
            We compare LOGIN with recent approaches~\cite{xu2017scene, zellers2018neural, yang2018graph, chen2019knowledge, tang2019learning, zhang2019graphical}. The results are summarized in~\tabref{tab:brc}.
            Here, although~\cite{xu2017scene, yang2018graph, chen2019knowledge} use the initial predicate representation as a union feature, they enable understanding of relational direction by incorporating contexts with iterative bipartite message passing, attentional graph convolution, and knowledge embedded routing, respectively.
            By using direction sensitive embedding and contextual information at the same time, LOGIN can outperform the recent methods by a large margin (6\% of mean performance gain compared to the state-of-the-art), implying that directional bias as well as contexts are crucial in recognizing direction.
            LOGIN is in a competitive position for VG dataset, which mainly contains uni-directional relationships, but significantly improves performance, especially for bidirectional relationships, which are common in the real world.
            
    \subsection{Quantitative Analysis}
    \label{section:quantitative}
        \paragraph{Model Ablations}
            We consider several ablations to investigate the importance of the major design choices in~\tabref{tab:ablation} (a). For clarity, we show the performance in the SGG task and the BRC task in a single table. 
            Exp 1 is the result of a vanilla version of LOGIN, \ie, \textsc{Baseline}, which shows the abysmal result, especially in the BRC setting. This means that the \textsc{Baseline} has no understanding of relational direction at all; thus, it can only predict symmetric relationships correctly.
            Exp 2 - Exp 5 examine the individual contributions of each model component. Especially, LIH (Exp2) and GIH (Exp4) have a significant impact on both SGG and BRC settings. It is noteworthy that contextual information (driven from GIH) also plays a key role in recognizing directions.
            We can see in Exp3 that DSE is relatively unremarkable in SGG settings, while it improves performance in BRC settings by a large margin.
            Although the unary effect of Attract \& Repel Loss $\mathcal{L}_{ar}$ (Exp 4) is not significant, using the loss with other components (Exp7) can further push the performance than without it (Exp 6).
            When all model components are combined (Exp 7), the model achieves the best performance in both SGG and BRC tasks, which implies that each component contains an orthogonal factor that complementarily boosts the performance.
            
        \paragraph{Optimal Variables}
            We conduct experiments to decide optimal variables of LOGIN in~\tabref{tab:ablation}
            The optimal feature extraction method is first investigated. Here, flattening the feature maintains richer information than pooling, thus shows the best results among the three choices: \textsc{avgpool}, \textsc{maxpool}, \textsc{flatten}.
            Then we examine the optimal number of layers in GIH: 4-layers produce the best results.
            Stacking multiple layers enables multi-hop communication, though it also increases the chance of introducing noisy information.
            On the other hand, stacking few layers cannot fully capture the higher-order contexts.
        
        \paragraph{Design Choices of DSE}
            In this experiment, we further explore the four design choices of Direction-Sensitive Encoding (DSE). Specifically, we investigate two approaches that have been adopted in most existing SGG literature -- (a) using only a union feature~\cite{xu2017scene, lu2016visual, dai2017detecting, yang2018graph, qi2019attentive, chen2019knowledge, chen2019counterfactual, li2017scene, li2018factorizable} (\textsc{Baseline}) and (b) fusing subject, object, and union without-permutations~\cite{zhang2017visual, yang2018shuffle, zellers2018neural, zhang2019graphical, tang2019learning, tang2020unbiased} -- and two variants of subject, object, and union fusion under the \textit{subject-precedes-object} constraint -- (c) sequential fusion and (d) parallel fusions (see~\figref{fig:fusion}).
            Except for the (a) among the four cases, the ordering of the subject and the object is fixed and therefore meets the directionality condition. Additionally, (c) and (d) consider the sum of all possible permutations. The difference between (c) and (d) is the order of fusion.
            We conduct experiments in two settings for performance comparison on fusion methods. The results are summarized in \tabref{tab:fusion}.
            Here, the performance difference between the four fusion methods in the SGC setting~(\tabref{tab:fusion} (a)) is not prominent, while the significance of combining three features is particularly evident in the BRC setting~(\tabref{tab:fusion} (b)), suggesting that union feature alone cannot give relational direction.
            In both settings, the use of permutations at the fusion phase showed better results than otherwise, and especially when fused in parallel, it showed the best results.

        \begin{table}[t!]
            \captionsetup{font=footnotesize}
            \caption{
                \textbf{Effectiveness of Graph Interaction Head (GIH)} compared to other graph neural networks (\eg, GCN~\cite{kipf2017semi}, GAT~\cite{velivckovic2018graph}).
            }
            \setlength{\tabcolsep}{10pt}
            \centering
            \begin{tabular}{@{}c@{}}
                \textrm{(a) Scene Graph Classification}
            \end{tabular}
            \resizebox{0.8\linewidth}{!}{
            \begin{tabular}{@{}ccccc@{}}
                \toprule[0.15em]
                \multicolumn{2}{r}{} & \multicolumn{3}{c}{\textit{SGCls}} \\ \cmidrule(l){3-5}
                \multicolumn{2}{c}{Methods} & R@20 & R@50 & R@100 \\ \midrule
                \multicolumn{2}{c}{\textsc{LOGIN /w GCN}~\cite{kipf2017semi}} & 33.8 & 37.5 & 39.7 \\ 
                \multicolumn{2}{c}{\textsc{LOGIN /w GAT}~\cite{velivckovic2018graph}} & 33.2 & 37.1 & 39.5 \\
                \rowcolor{Gray}
                \multicolumn{2}{c}{\textbf{\textsc{LOGIN /w GIH} (Ours)}} & \textbf{34.5} & \textbf{38.8} & \textbf{40.5} \\
                \bottomrule[0.15em]
            \end{tabular}
            }
            \begin{tabular}{@{}c@{}}
                \\
                \textrm{(b) Bidirectional Relationship Classification}
            \end{tabular}
            \resizebox{\linewidth}{!}{
            \begin{tabular}{@{}cccccc@{}}
                \toprule[0.15em]
                \multicolumn{2}{r}{} & \multicolumn{4}{c}{\textit{PredCls}} \\ \cmidrule(l){3-6}
                \multicolumn{2}{c}{Fusion} & pR@2 & pR@4 & pR@8 & pR@16 \\ \midrule
                \multicolumn{2}{c}{\textsc{LOGIN /w GCN}~\cite{kipf2017semi}} & 7.8 & 12.4 & 16.5 & 20.5 \\
                \multicolumn{2}{c}{\textsc{LOGIN /w GAT}~\cite{velivckovic2018graph}} & 7.9 & 12.3 & 16.3 & 20.3 \\
                \rowcolor{Gray}
                \multicolumn{2}{c}{\textbf{\textsc{LOGIN /w GIH} (Ours)}} & \textbf{8.6} & \textbf{13.1} & \textbf{17.6} & \textbf{21.1} \\
                \bottomrule[0.15em]
            \end{tabular}
            }
            \label{tab:gih}
        \end{table}

        \paragraph{Effectiveness of GIH}
            We examine the effectiveness of GIH by comparing GIH with two representative message passing graph neural networks in~\tabref{tab:gih}: Graph Convolutional Network (GCN)~\cite{kipf2017semi} and Graph Attention Network (GAT)~\cite{velivckovic2018graph}.
            GCN aggregates feature information via a non-euclidean convolution operation from a node’s neighborhood.
            As opposed to GCNs, GAT allows for implicitly assigning different importances to nodes of a same neighborhood, enabling a leap in model capacity.
            Unlike them, layer-wise propagation rule of GIH considers not only nodes but also edges as a neighborhood, allowing the model to leverage higher-order contexts for node update.
            From the results, we see that GAT does not improve the performance upon the GCN.
            The results demonstrate the effectiveness of GIH in predicting both object and relationships categories (Scene Graph Classification). Since LOGIN equipped with GIH exploits richer information (\eg, edge), it is also strong in understanding relational direction (Bidirectional Relationship Classification).

        \begin{table}[t!]
            \captionsetup{font=footnotesize}
            \caption{
                \textbf{Per-type predicate classification results.} Only top-20 frequent predicates are shown. The evaluation metric is \textit{R@50}.
            }
            \centering
            \resizebox{\linewidth}{!}{
            \begin{tabular}{@{}ccc|ccc@{}}
                \toprule[0.15em]
                predicate & Baseline & LOGIN & prediate & Baseline & LOGIN \\
                \midrule
                on & 66.3 & 88.1 & sitting on & 32.2 & 61.0 \\
                has & 47.7 & 87.5 & under & 35.9 & 52.5 \\
                wearing & 68.9 & 93.7 & riding & 26.3 & 83.0 \\
                of & 42.8 & 82.4 & in front of & 8.9 & 29.4 \\ 
                in & 47.7 & 64.1 & standing on & 16.7 & 37.7 \\
                near & 19.4 & 52.3 & at & 39.5 & 57.8 \\
                with & 18.1 & 45.9 & attached to & 12.1 & 21.0 \\
                behind & 20.7 & 57.0 & carrying & 23.9 & 62.1 \\
                holding & 31.4 & 78.7 & walking on & 10.5 & 59.3 \\
                above & 18.2 & 51.2 & over & 9.5 & 28.4 \\
                \bottomrule[0.15em]
            \end{tabular}
            }
            \label{tab:top20}
        \end{table}

        \paragraph{Per-type Predicate Recall}
            We expect the model to better understand each predicate by allowing attention mechanism of LIH to capture the predicate label-specific representation well and Attract \& Repel Loss to help separate inter-class and aggregate intra-class predicates in the embedding space. In order to ensure that the proposed model solves the ambiguity issue well, we compare our LOGIN with Baseline under the Recall@50 metric for the top-20 frequent predicates in~\tabref{tab:top20}.
            Compared to Baseline, we observe a significant performance improvement in all predicate classes.
            Specifically, our system better understands the geometric predicate (\eg., on, in front of, behind, above, under), possessive predicates (\eg, has, of, wearing), and semantic predicates (\eg, holding, walking).
            This suggests that explicit separation on predicate embedding space properly solves the ambiguity problem.
            
    \subsection{Qualitative Analysis}
    \label{section:qualitative}
        \begin{figure*}[p!]
            \centering5
            \includegraphics[width=\linewidth]{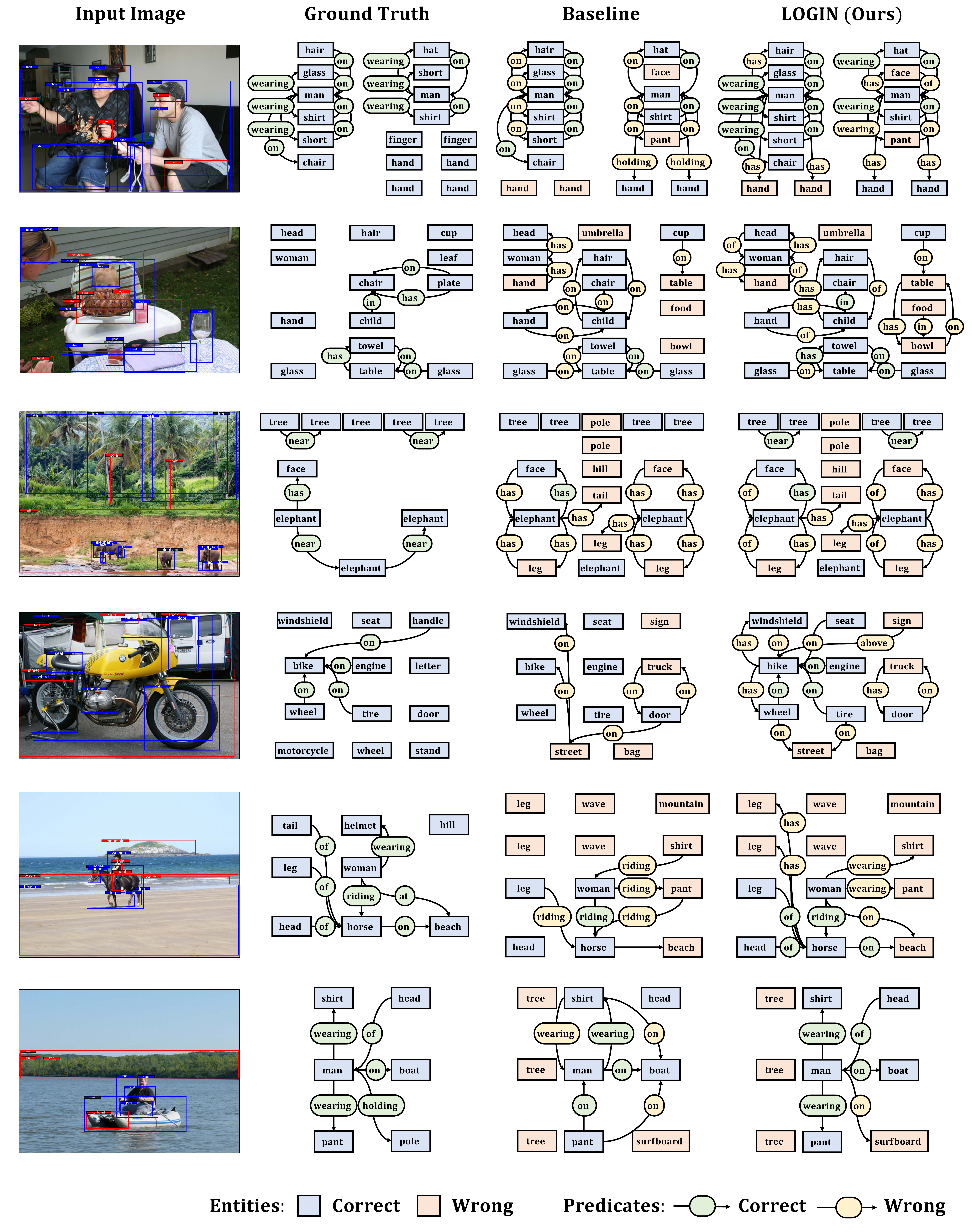}
            \captionsetup{font=footnotesize}
            \caption{
                \textbf{Qualitative examples.} The first column shows input images with entity proposals. From the second to fourth columns, we show the scene graphs of ground-truth, \textsc{Baseline}, and \textsc{LOGIN} respectively. The bounding boxes or nodes are colored in either blue (correct) or red (wrong). The predicates are colored in either green (correct) or yellow (wrong). Examples of the first two rows contain bidirectional relationships, but not the rest. We see that LOGIN produces more diverse predicates and can successfully distinguish asymmetric relationships while \textsc{Baseline} model fails.
            }
            \label{fig:QualitativeResults}
        \end{figure*}
        
        To better see how LOGIN understands the relational direction, we provide qualitative examples in~\figref{fig:QualitativeResults}.
        Here, We compare the result of \textsc{Baseline} model and \textsc{LOGIN} with the corresponding ground-truth scene graph.
        As we can see in the results of first two rows, \textsc{Baseline} model produces the same result for a pair of entities regardless of direction. What is worse is that the whole scene graphs use almost the same predicates for defining relationships. In other words, the \textsc{Baseline} model neither considers relational-direction nor lexical diversity.
        On the other hand, LOGIN can successfully identify relational direction, thanks to the embedded direction-awareness, and it is also more diverse in terms of vocabulary.
        More interestingly, even though predictions of LOGIN are not matched, the results are seemingly plausible. For example, in the third row, detected \texttt{tail}, \texttt{legs}, and \texttt{face} of an \texttt{elephant} are false positives in terms of ground-truth, but they seem to be correct in reality. Also, relationships associated with false positives are somewhat reasonable (\eg, $\texttt{elephant} \xrightarrow{\texttt{has}} \texttt{leg}$, $\texttt{leg} \xrightarrow{\texttt{of}} \texttt{elephant}$).
        

%% file: 6.conclusion.tex
\section{Conclusion}
    This paper discusses three fundamental challenges in SGG task: 1) Ambiguity, 2) Asymmetry, and 3) Higher-order contexts. Motivated by the analysis and to tackle the issues effectively, we present a new unified framework, LOGIN. Our framework enables predicting the scene graph in a local-to-global and bottom-to-up manner, leveraging the possible complementariness effectively. We achieved state-of-the-art on Visual Genome benchmark.
    Last but not least, we present a new diagnostic task called Bidirectional Relationship Classification (BRC) and observe that our method outperforms competing methods significantly.

%% file: 7.references.tex
\ifCLASSOPTIONcaptionsoff
  \newpage
\fi

{\footnotesize 
\bibliographystyle{IEEEtran}
\bibliography{ieee_tnnls}
}

%% file: 8.biography.tex
\begin{IEEEbiography}[{\includegraphics[width=1in,height=1.25in,clip,keepaspectratio]{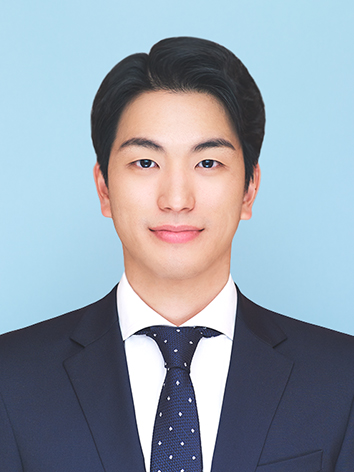}}]{Sangmin Woo} (Student Member, IEEE)
    is currently pursuing the Ph.D. degree in electrical engineering at Korea Advanced Institute of Science and Technology (KAIST), Daejeon, Korea. He received an M.S. degree in Electrical Engineering and Computer Science from Gwangju Institute of Science and Technology (GIST), Gwangju, Korea, in 2021, and a B.S. degree in Electrical Engineering from Kyungpook National University, Daegu, Korea, in 2019. His research interests lie in computer vision and machine learning, especially in a high-level visual understanding.
\end{IEEEbiography}

\begin{IEEEbiography}[{\includegraphics[width=1in,height=1.25in,clip,keepaspectratio]{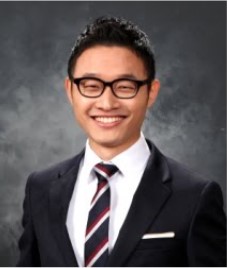}}]{Junhyug Noh} (Member, IEEE)
    is a postdoctoral researcher at Lawrence Livermore National Laboratory (LLNL). He received the B.S. in Computer Science and Engineering \& Statistics from Seoul National University in 2013, and the M.S. and Ph.D. in Computer Science Engineering from Seoul National University in 2015 and 2020, respectively. His research has focused on artificial intelligence, machine learning, and computer vision with a particular interest in object detection and its related high-level vision tasks such as semantic/instance segmentation, scene understanding, and image captioning.
\end{IEEEbiography}

\begin{IEEEbiography}[{\includegraphics[width=1in,height=1.25in,clip,keepaspectratio]{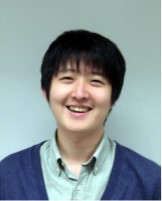}}]{Kangil Kim} (Member, IEEE)
    received the B.S. degree in computer science from the Korea Advanced Institute of Science and Technology, Daejeon, South Korea, in 2006, and the Ph.D. degree from Seoul National University, Seoul, South Korea, in 2012. He was a Senior Researcher with the Natural Language Processing Group, Electronics and Telecommunications Research Institute, Seoul, until 2016, and an Assistant Professor with the Computer Science and Engineering Department, Konkuk University, until 2019. He is currently an Assistant Professor with the Electronics Engineering and Computer Science Department and Artificial Intelligence Graduate School, Gwangju Institute of Science and Technology, Gwangju, South Korea. His research interests include artificial intelligence, evolutionary computation, machine learning, and natural language processing.
\end{IEEEbiography}